\documentclass{article}

\usepackage[preprint]{neurips_2025}

\usepackage[utf8]{inputenc} %
\usepackage[T1]{fontenc}    %
\usepackage{hyperref}       %
\usepackage{url}            %
\usepackage{booktabs}       %
\usepackage{amsfonts}       %
\usepackage{nicefrac}       %
\usepackage{microtype}      %
\usepackage{xcolor}         %
\usepackage{xcolor,colortbl} 
\usepackage{adjustbox}%

\usepackage{verbatim}

\usepackage{multirow}
\usepackage{graphicx}
\usepackage{enumitem}
\usepackage{caption}

\usepackage{multirow}
\usepackage{tabularx}
\newcolumntype{Y}{>{\centering\arraybackslash}X}
\newcolumntype{C}[1]{>{\centering}p{#1}}
\newcolumntype{Z}{>{\raggedleft\arraybackslash}X}

\newcommand{\STAB}[1]{\begin{tabular}{@{}c@{}}#1\end{tabular}}

\usepackage[outline]{contour}
\newcommand{\uparrowaligned}{\raisebox{0.15em}{\scalebox{0.75}{\contour{black}{$\uparrow$}}}}
\newcommand{\downarrowaligned}{\raisebox{0.15em}{\scalebox{0.75}{\contour{black}{$\downarrow$}}}}

\setcitestyle{square}
\setcitestyle{numbers}

\usepackage{xspace}
\makeatletter
\DeclareRobustCommand\onedot{\futurelet\@let@token\@onedot}
\def\@onedot{\ifx\@let@token.\else.\null\fi\xspace}

\def\eg{\emph{e.g}\onedot}

\makeatother

\newcommand{\cellbest}{\cellcolor{red!23}}
\newcommand{\cellsecond}{\cellcolor{orange!26}}
\newcommand{\cellthird}{\cellcolor{yellow!29}}

\usepackage{multibib}
\newcites{supp}{Supplementary References}

\newcommand{\name}{ViDAR}

\newcommand{\authorskip}{\qquad}

\title{ViDAR: Video Diffusion-Aware\\4D Reconstruction From Monocular Inputs}

\author{%
  Michal Nazarczuk$^1$\thanks{equal contribution} \authorskip Sibi Catley-Chandar$^{1,2}${\footnotemark[1]} \authorskip Thomas Tanay$^1$\\ \textbf{Zhensong Zhang}$^1$ \authorskip \textbf{Gregory Slabaugh}$^2$ \authorskip \textbf{Eduardo Pérez-Pellitero}$^1$\\[2mm]
  $^1$ Huawei Noah's Ark Lab \qquad $^2$ Queen Mary University of London
}

\begin{document}

\maketitle

\begin{figure}[h]
\vspace{-2em}
    \centering
    \includegraphics[width=0.8\linewidth]{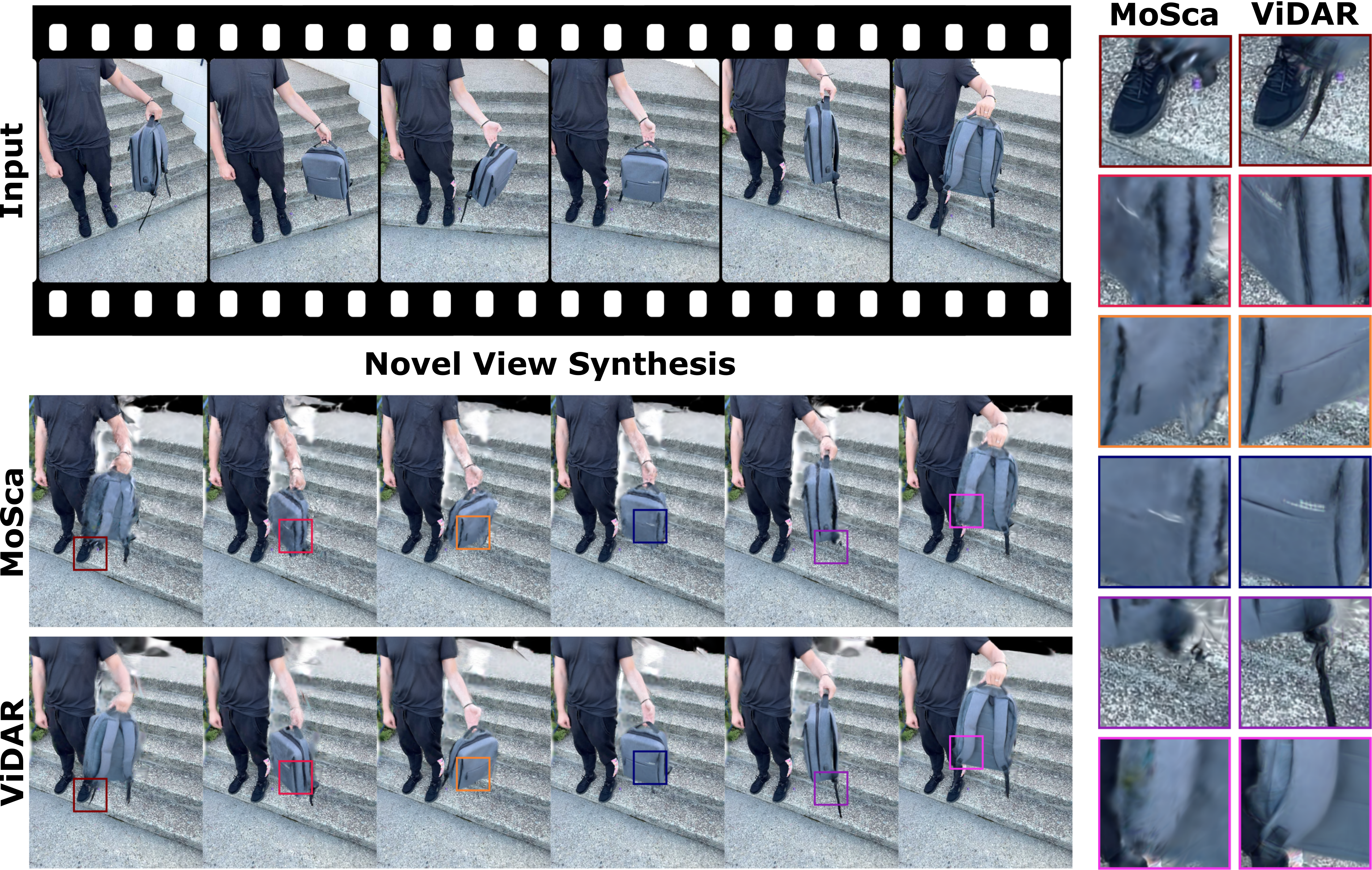}
    \caption{\name{} provides a novel framework for Monocular Novel View Synthesis utilising a diffusion-aware reconstruction framework.}
    \label{fig:teaser}
\end{figure}

\begin{abstract}
Dynamic Novel View Synthesis aims to generate photorealistic views of moving subjects from arbitrary viewpoints. This task is particularly challenging when relying on monocular video, where disentangling structure from motion is ill-posed and supervision is scarce. We introduce Video Diffusion-Aware Reconstruction (ViDAR), a novel 4D reconstruction framework that leverages personalised diffusion models to synthesise a pseudo multi-view supervision signal for training a Gaussian splatting representation. By conditioning on scene-specific features, ViDAR recovers fine-grained appearance details while mitigating artefacts introduced by monocular ambiguity. To address the spatio-temporal inconsistency of diffusion-based supervision, we propose a diffusion-aware loss function and a camera pose optimisation strategy that aligns synthetic views with the underlying scene geometry. Experiments on DyCheck, a challenging benchmark with extreme viewpoint variation, show that ViDAR outperforms all state-of-the-art baselines in visual quality and geometric consistency. We further highlight ViDAR’s strong improvement over baselines on dynamic regions and provide a new benchmark to compare performance in reconstructing motion-rich parts of the scene. Project page: \url{https://vidar-4d.github.io/}.
\end{abstract}

\section{Introduction}

4D reconstruction from monocular inputs is a challenging problem where the goal is to recover a 3D representation of a dynamic scene. It is increasingly important for modelling, comprehending, and interacting with the physical world and supports a wide range of downstream applications, ranging from augmented reality to generating data for training robust AI models~\cite{yang2025novel}.

Casually captured monocular videos are ubiquitous, however reconstructing 3D structure from them remains an inherently ill-posed problem. Static regions of the scene can typically be reconstructed well due to effective multi-view capture \cite{gao2022}. However for dynamic regions, depth information is not directly observable from a single viewpoint; in other words, it is difficult to disentangle the motion of the camera from motion within the scene. 
To mitigate this ambiguity many existing approaches impose strong regularisation \cite{huang2024, park2025, yang2024} in the form of geometric assumptions, such as the object's rigidity, that constrains the dynamics of the scene. Others \cite{lei2024, liu2025, wang2024, zhou2023, zhu2024} leverage learned priors, particularly those derived from large-scale models (\eg monocular depth), to guide the reconstruction. 
While regularization based methods  \cite{lei2024, wang2024} achieve geometrically compact scene representations, they often fall short in rendering high-quality, photorealistic appearances. Conversely recent generative approaches utilise powerful diffusion models to achieve higher visual quality in tasks such as single image to 3D \cite{lin2025, liu2024a, liu2024, xu2024a, xu2024, zou2024} and monocular reconstruction \cite{wu2024b} but struggle to maintain spatio-temporal coherence, limiting their applicability in scenarios that demand accurate spatial reconstruction and temporal consistency, particularly in dynamic, real-world settings. %

To tackle these challenges, we present Video Diffusion-Aware 4D Reconstruction (\name{}), a monocular video reconstruction approach that leverages diffusion models as powerful appearance priors through a novel diffusion-aware reconstruction framework, which allows for improving visual fidelity without the loss of spatio-temporal consistency. 
We first train a monocular reconstruction baseline and generate a set of typically degraded multi-view images by sampling diverse camera poses and rendering the novel viewpoints. We then adopt a DreamBooth-style personalisation strategy~\cite{ruiz2023}, and tailor a pretrained diffusion model to the input video, which we use as a generative enhancer to inject rich visual information back into the degraded renders. This effectively generates a set of high-fidelity pseudo-multi-view observations for our scene, although due to the nature of the diffusion process, the resulting images are not necessarily spatially consistent. We observe that naively using these views as supervision leads to reconstructions degraded by artefacts and geometric inconsistencies. To mitigate this, we propose a method of diffusion-aware reconstruction, which selectively applies diffusion-based guidance to dynamic regions of the scene while jointly optimising the camera poses associated with the diffused views.

To the best of our knowledge, \name{} is the first approach to incorporate a diffusion prior into monocular video reconstruction in a geometrically consistent manner. We demonstrate substantially improved qualitative and quantitative results compared to existing techniques (see Tabs.\,\ref{tab:results-covis}, \ref{tab:results-dynamic}, Figs.\,\ref{fig:teaser}, \ref{fig:quali-1}), highlighting the effectiveness of diffusion-guided supervision when integrated with a reconstruction pipeline that accounts for geometric consistency.

We summarise our contributions as follows:
\begin{enumerate}[topsep=-2pt,itemsep=0pt]
    \item A personalised diffusion enhancement strategy that improves appearance quality by refining newly sampled renderings using a DreamBooth-adapted model.
    \item A diffusion-aware reconstruction framework that combines dynamic-region-focused diffusion guidance with joint optimisation of the sampled camera poses for geometrically consistent reconstruction.
    \item An extensive experimental evaluation, including both quantitative and qualitative comparisons with prior work, the introduction of a dynamic-region specific benchmark, as well as ablation studies isolating the impact of each component. 
\end{enumerate}

\section{Related work}

\paragraph{4D reconstruction} Advances in novel view synthesis include the introduction of two seminal reconstruction paradigms, namely Neural Radiance Fields (NeRF)\,\cite{mildenhall2020} and 3D Gaussian Splatting (3DGS)\,\cite{kerbl2023}. These developments in static scene reconstruction were quickly followed by several works on dynamic content. NeRF-based methods for video reconstruction include D-NeRF\,\cite{pumarola2020}, StreamRF\,\cite{li2022b}, HexPlane\,\cite{cao2023}, K-Planes\,\cite{fridovich2023}, Tensor4D\,\cite{shao2023}, MixVoxels~\cite{wang2023}.
Similarly, Gaussian Splatting developments enabled research on dynamic novel view synthesis. Multi-view videos were reconstructed by: GaussianFlow\,\cite{lin2024}, 4DGS\,\cite{wu2024}, STG\,\cite{li2024}, SWinGS\,\cite{shaw2024}, Ex4DGS\,\cite{lee2024}.

\paragraph{Monocular reconstruction} The task of 4D monocular video reconstruction can be seen as a special case of 4D reconstruction under substantially more challenging conditions. This is due to the problem often being ill-posed: many of the target object surfaces may be seen only from one viewpoint throughout the video. Notably, among NeRF-based approaches, NSFF\,\cite{li2021} proposes a time varying flow field, whereas Nerfies\,\cite{park2021b}, HyperNeRF\,\cite{park2021a}, DyCheck (T-NeRF)\,\cite{gao2022}, DyBluRF\,\cite{bui2023}, RoDynRF\,\cite{liu2023}, CTNeRF\,\cite{miao2024} use a canonical representation with a time-dependent deformation. DynIBaR\,\cite{li2023} uses Image Based Rendering for reconstruction.
With Gaussian Splatting advancements, Dynamic 3D Gaussians\,\cite{luiten2023} learn explicit motion of every Gaussian, whereas 4DGS\,\cite{wu2024}, Deformable 3DGS\,\cite{yang2024}, SC-GS\,\cite{huang2024} use a deformation field for transformation from canonical space. SplineGS\,\cite{park2025} constrains the motion of Gaussians to splines to ensure temporal smoothness. DynPoint\,\cite{zhou2023} and MotionGS\,\cite{zhu2024} use an optical flow estimator for additional supervision. PGDVS\,\cite{zhao2024} and BTimer\,\cite{liang2024} propose a transformer-based approach for generalisable reconstruction. Dynamic Gaussian Marbles \cite{stearns2024} adopt a divide-and-conquer strategy to merge sets of Gaussians and create long trajectories, and restrict representation to isotropic Gaussians. MoDGS\,\cite{liu2025} improves the supervision from depth priors. D-NPC\,\cite{kappel2025} proposes the use of neural implicit point cloud as the representation for monocular reconstruction.
MoSca\,\cite{lei2024} and Shape of Motion\,\cite{wang2024} both utilise priors from pretrained foundational models (depth, optical flow, 2D tracking). Similarly, they both reconstruct static and dynamic content separately, and describe the motion of the Gaussians with lower dimensionality basis functions.

\paragraph{Diffusion enhanced reconstruction} Several recent approaches explore the use of diffusion models to guide the reconstruction. ReconFusion\,\cite{wu2024c} trains a diffusion model on a set of object images, and uses it to score the quality of sparse reconstruction, guiding it with RGB loss. DpDy\,\cite{wang2024b} uses Score Distillation Sampling (SDS)\,\cite{poole2023} to supervise reconstruction with the use of image and depth diffusion model. CAT4D\,\cite{wu2024b}, concurrent to our work, uses a video-diffusion model to generate additional static cameras for the input video, followed by the reconstruction process. MVGD\,\cite{guizilini2025} proposes a direct rendering of novel views and depth as a conditional generative task. Other diffusion-based approaches include text or a single image to 3D generation\,\cite{li2024a, liang2024b, lin2025, liu2024a, liu2024, shriram2025, tang2024, wimmer2025, xu2024a, xu2024, yang2024a, yi2024, yu2024, zeng2024, zou2024}. Notably, our approach uses a monocular video as an input, and uses a personalised diffusion model along with our diffusion-aware reconstruction for accurate geometry modelling.

\section{Method} \label{sec:method}

\begin{figure}
    \centering
    \includegraphics[width=1.0\linewidth]{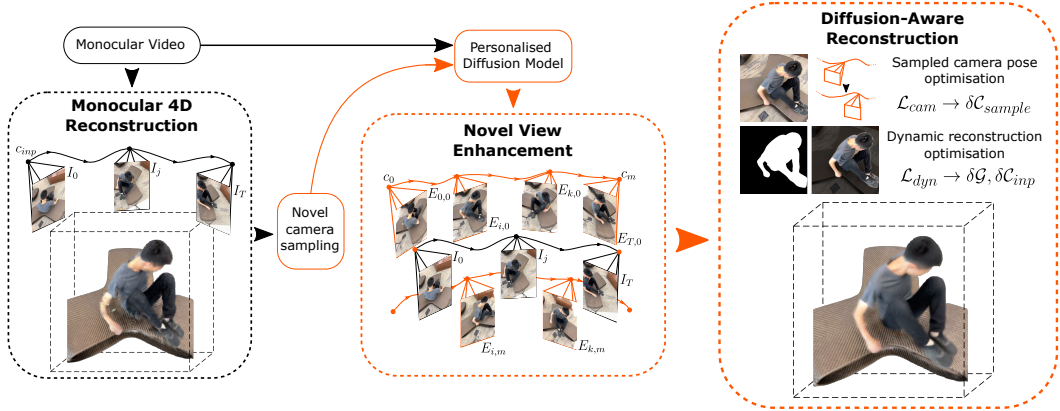}
    \caption{A high-level overview of \name{}. The input video is used to create a 4D reconstruction with a monocular approach. Further, novel camera views are sampled and enhanced with a personalised diffusion model for each scene. This constitutes a set of pseudo-multi-view supervision examples. Finally, our approach optimises the 4D representation with the use of original video and new multi-view cues, in a diffusion-aware manner.}
    \label{fig:main}
    \vspace{-1em}
\end{figure}

Our method incorporates several stages which can be seen in Figure\,\ref{fig:main}. Firstly, we use a monocular reconstruction baseline to obtain a 4D representation of the scene (Sec.\,\ref{sec:mono_recon}), and generate a set of degraded multi-view renders from sampled novel camera poses. Next, we personalise a diffusion model using the input video (Sec.\,\ref{sec:diffusion}), which is used to enhance the degraded renders (Sec.\,\ref{sec:sampling_enhancement}). Finally, we use the new set of enhanced pseudo-multi-view images to supervise and refine the 4D representation of the scene (Sec.\,\ref{sec:pseudo_multi_recon}), in a diffusion-aware manner.

\subsection{Monocular Reconstruction} \label{sec:mono_recon}

Given a casual monocular video of a dynamic scene with $T$ frames $\mathcal{I} = [I_1, I_2, \dots I_T]$, we perform initial reconstruction of the scene using an off-the-shelf 4D monocular reconstruction method, specifically, we use MoSca \cite{lei2024} in our implementation. The method reconstructs two sets of Gaussians for the given scene, namely static Gaussians $\mathcal{G}_s$, and dynamic Gaussians $\mathcal{G}_d$, that together create the scene representation: $\mathcal{G} = \mathcal{G}_d \cup \mathcal{G}_s$. MoSca leverages several priors in the reconstruction process: depth, optical flow, 2D tracking. Firstly, the optical flow is used to estimate the epipolar error and to determine the likelihood of image regions belonging to dynamic or static content. This is followed by the joint reconstruction of the static part of the scene  $\mathcal{G}_s$ and fine-tuning of the input camera pose $c_{inp}$. With that, a \textit{scaffold}, a low dimensionality motion representation, is built through lifting 2D tracklets belonging to dynamic regions into 3D using depth information. Finally, a photometric reconstruction is performed to optimise the scene $\mathcal{G}$, enabling rendering of novel views $R$ of the scene.

\subsubsection{Track Anything Gaussian Classification} \label{sec:ta_masks}

We note that the epipolar error analysis introduced by MoSca for classification of dynamic parts of the image leads to occurrences of floater artefacts due to the inclusion of background among dynamic Gaussians. This may not be reflected heavily in quantitative performance, but leads to a decrease in the quality of generated pseudo-multi-view samples (Sec.\,\ref{sec:sampling_enhancement}). To improve the constraint on dynamic Gaussians' locations, we use dynamic masks ${D}_t$ obtained from Track Anything\,\cite{yang2023} to reconstruct the static part of the scene  $\mathcal{G}_s$ and generate motion scaffolds (as in MoSca\,\cite{lei2024}). 

\subsection{Diffusion Enhancement} \label{sec:diffusion}

We utilise a Stable Diffusion\,\citep{rombach2021} model, specifically the pretrained Stable Diffusion XL (SDXL)\,\citep{podell2024} to improve the quality of rendered images and guide the reconstruction process. Following the observations of ReconFusion\,\citep{wu2024c}, we decide to use a multistep denoising process, in contrast to Score Distillation Sampling\,\citep{poole2023}. Conversely, given a sampled image $R_{m, t}$ from camera $c_m$ at the time $t$, we follow a standard text-to-image\,\citep{rombach2021} process and encode the image into latent space: $x_0=\mathcal{E}\left(R_{m, t}\right)$. Further, instead of generating the noisy latent for image generation, we introduce $k$ steps of noise into the image-sourced latent $x_0 \rightarrow x_k$ using the original noise scheduler (here, Discrete Euler\,\citep{karras2022}). We then follow the denoising process for $k$ steps to achieve a denoised latent $\hat{x}_0$ which is then decoded to an enhanced version of the input image: $E_{m, t}=\mathcal{D}\left(\hat{x}_0\right)$.

\paragraph{Personalisation}
Similarly to some of the recent reconstruction approaches, \eg \citet{wang2024b}, we apply the Dreambooth \citep{ruiz2023} fine-tuning approach to the SDXL model. To this end, we treat an input video $\mathcal{I}$ as a collection of images and fine-tune the diffusion model for a given scene such that a specific text token triggers the model to follow the appearance of the scene.

\subsubsection{View Sampling and Rendering Enhancement} \label{sec:sampling_enhancement}

Given the scene-personalised diffusion model, we utilise the previously trained monocular reconstruction to generate a set of pseudo-multi-view ground truth images. Firstly, we sample $M$ sets of images for each timestep $t \in [0, T]$, effectively adding $M$ new cameras with parameters $c_m$ where $m\in[0,M]$ and $c_m \in \mathcal{C}_{sample}$, where $\mathcal{C}_{sample}$ constitutes a set of new camera trajectories. To this end, we select two existing views (as a camera position and rotation), a random one, and a challenging view (with the furthest distance from the mean) and sample a new view as their weighted linear combination. To introduce variety in the difficulty of the sampled views, we gradually increase the blending weight of the views towards the most challenging ones from the input trajectory. Simultaneously, we introduce noise of an increasing amplitude to the new cameras.

Thereafter, we use our trained monocular reconstruction to render a set of $M$ new camera views $R_{m,t}$ for each timestep. Further, we use our personalised diffusion model to enhance the rendered images  $R_{m,t}\rightarrow E_{m,t}$. This constitutes a new set of supervision images in a multi-view setting. We have chosen to generate a whole multi-view dataset in a single step instead of performing the enhancement on-the-fly. This enables the samples to be reused and reduces the computational demands (especially on GPU memory).

\subsection{Diffusion-Aware Reconstruction} \label{sec:pseudo_multi_recon}

We use our generated dataset $\{E_{m,t}\}$ as additional supervision to re-train our 4D monocular reconstruction method to predict a higher quality output $\hat I_{m,t}$. However using these sampled views for training is challenging. The outputs $E_{m,t}$ of our personalised video diffusion models are high-fidelity and also preserve structure and coarse geometry, but due to the nature of the diffusion process and random noise schedule, they are not spatio-temporally consistent at the level of fine-grained detail and texture. This manifests as flickering and shifts of textures between consecutive frames. In some cases, coarse geometry may also be hallucinated, e.g. in novel viewpoints not seen during training. If we naively used these outputs to supervise monocular 4D reconstruction, the lack of spatio-temporal consistency in the training data would cause the model to either converge to a mean radiance value and cause blurry renderings, or to overfit to individual frames and learn a temporally inconsistent reconstruction. We propose the following mechanisms to overcome these challenges. 

\subsubsection{Dynamic Reconstruction} \label{sec:dyn_recon}
While dynamic regions of a scene are under-observed, static regions may be captured from multiple viewpoints across time, effectively creating multi-view supervision. Hence supervision is unnecessary in static regions and in fact could cause the quality to reduce, particularly if spatially inconsistent.  We compute a mask of the dynamic regions of the scene ${D}_{m,t}$ using Track Anything \citep{yang2023}, and apply this mask to our data to mask out the static regions $ {{E}_{m,t}^{dyn}} = {E}_{m,t} \odot {D}_{m,t}$, where $\odot$ denotes element-wise multiplication, and also to our predicted output  ${\hat{I}_{m,t}^{dyn}} = \hat{I}_{m,t} \odot {D}_{m,t}$. This ensures that only the dynamic regions of the scene are supervised by our generated data, which reduces the convergence to the mean effect in the static reconstruction and reduces floaters. For dynamic regions, we introduce a perceptual loss \cite{zhang2018} to encourage our reconstruction to be texturally rich and reduce blur caused by training on spatially misaligned psuedo-GTs. During training we compute the loss as $\mathcal{L}_{dyn} = |{E}_{m,t}^{dyn} - \hat{I}_{m,t}^{dyn}|_{1}+ \lambda_{p} |{E}_{m,t}^{dyn} - \hat{I}_{m,t}^{dyn}|_{vgg} + \lambda_{s} |{E}_{m,t}^{dyn} - \hat{I}_{m,t}^{dyn}|_{ssim}$, where $\left|\cdot\right|_{1}$
is the L1 loss, $\left|\cdot\right|_{vgg}$ is the perceptual loss using a pretrained VGG network \cite{simonyan15}, $\left|\cdot\right|_{ssim}$ is the SSIM \cite{wang2004} loss and $\lambda_{p}$ and $\lambda_{s}$ are hyperparameters set to 0.1. The dynamic loss, $\mathcal{L}_{dyn}$, is applied in addition to the default losses from the monocular reconstruction method, and is backpropagated to update $\mathcal{G}$ and  $\mathcal{C}_{inp}$.

\subsubsection{Sampled Camera Pose Optimisation} \label{sec:cam_opti}

Camera poses of casually captured monocular videos are typically noisy due to the difficulty of disentangling scene motion from camera motion, thus the need to optimise $\mathcal{C}_{inp}$ in many monocular reconstruction methods \cite{lei2024, wang2024}. Our sampled camera poses $\mathcal{C}_{sample}$ are interpolated from $\mathcal{C}_{inp}$ and so are also noisy. As our psuedo-GTs corresponding to $\mathcal{C}_{sample}$ are not always spatially consistent, it is even more difficult to disentangle scene motion from camera motion.  To compensate for this, it is necessary to optimise our sampled camera poses during training to ensure the psuedo-GTs are aligned with the underlying scene geometry. However unlike dynamic reconstruction (Sec.~\ref{sec:dyn_recon}) where we only use the dynamic masked region for supervision, we use the entire image ${E}_{t,m}$ as supervision for sampled camera pose optimisation. Despite fine-grained textural flickering, the coarse structure present in static regions provides a more consistent supervision signal for localisation than using only dynamic regions. We compute the loss as $\mathcal{L}_{cam} = |{E}_{m,t} - \hat{I}_{m,t}|_{1}+ \lambda_{p} |{E}_{m,t} - \hat{I}_{m,t}|_{vgg} + \lambda_{s} |{E}_{m,t} - \hat{I}_{m,t}|_{ssim}$. The camera loss, $\mathcal{L}_{cam}$, is backpropagated separately to other losses and only updates $\mathcal{C}_{sample}$.

\section{Results}

\begin{table}

\centering
\caption{Quantitative results on co-visibility masked regions of scenes from the DyCheck (iPhone) dataset. Best, second and third results are highlighted in red, orange and yellow respectively.   SoM-5 is full-res with wheel and space-out excluded.}
\begin{tabularx}{\linewidth}{llZZZ}  
    \toprule
     & Method & PSNR-m\,\uparrowaligned & SSIM-m\,\uparrowaligned & LPIPS-m\,\downarrowaligned \\ 
      \midrule
      \multirow{14}{*}{\STAB{\rotatebox[origin=c]{90}{Half-Res}}} & T-NeRF\,\citep{gao2022} & 16.96 & 0.5772 & 0.3789\\ 
        & NSFF\,\citep{li2021} & 15.46 & 0.5510 & 0.3960 \\ 
        & Nerfies\,\citep{park2021b} & 16.45 & 0.5699 & 0.3389 \\ 
        & HyperNeRF\,\citep{park2021a} & 16.81 & 0.5693 & 0.3319\\ 
        & 4DGS\,\citep{wu2024} & 13.64 & - & 0.4280\\
        & PGDVS\,\citep{zhao2024} & 15.88 & 0.5480 & 0.3400 \\
        & DynPoint\,\citep{zhou2023} & 16.89 & 0.5730 & -\\
        & DyBluRF\,\citep{bui2023} & 17.37 & 0.5910 & 0.3730\\ 
        & D-NPC\,\citep{kappel2025} & 16.41 & 0.5820 & 0.3190 \\ 
        & RoDynRF\,\citep{liu2023} & 17.10 & 0.5340 & 0.5170\\ 
        & Gaussian Marbles\,\citep{stearns2024} & 16.02 & 0.5416 & 0.3398\\ 
        & SoM\,\citep{wang2024} & \cellthird18.62 &\cellthird 0.6820 & \cellsecond0.2382\\
        & MoSca\,\citep{lei2024} & \cellsecond19.32 & \cellsecond0.7060 & \cellthird0.2640 \\
        & Ours & \cellbest19.69 & \cellbest0.7126 & \cellbest0.2231\\
    \midrule
      \multirow{4}{*}{\STAB{\rotatebox[origin=c]{90}{Full-Res}}} &  Gaussian Marbles\,\citep{stearns2024} &  15.84 & 0.5434 & 0.5681 \\  
      & SoM\,\citep{wang2024}  & \cellthird17.98 & \cellthird0.6422 & \cellsecond0.3718 \\ 
    &MoSca\,\citep{lei2024} &  \cellsecond18.44 & \cellsecond0.6560 &  \cellthird0.4193 \\
    &Ours & \cellbest19.00 & \cellbest0.6672 & \cellbest0.3623 \\
    \midrule
      \multirow{9}{*}{\STAB{\rotatebox[origin=c]{90}{SoM-5}}} &  T-NeRF\,\citep{gao2022} &15.60 & 0.5500 & 0.5500 \\ 
        & HyperNeRF\,\citep{park2021a}  & 15.99 & 0.5900 & 0.5100\\ 
        & DynIBaR\,\citep{li2023}  & 13.41 & 0.4800 & 0.5500 \\
        & Gaussian Marbles\,\citep{stearns2024} & 16.03 & 0.5425 & 0.5807 \\
        & SoM\,\citep{wang2024}  &16.72 & \cellthird0.6300  & 0.4500 \\ 
        & CAT4D\,\citep{wu2024b}  & \cellthird17.39 & 0.6070 & \cellbest0.3410\\
        & MoSca\,\citep{lei2024} &  \cellsecond18.34 & \cellsecond0.6636 & \cellthird0.4321 \\ 
        & Ours &  \cellbest18.76 & \cellbest0.6751 & \cellsecond0.3774 \\
    \bottomrule
\end{tabularx}
\label{tab:results-covis}
\end{table}

\subsection{Datasets}

We evaluate the performance of \name~on the DyCheck dataset \citep{gao2022}. DyCheck was introduced as a real world benchmark for evaluating monocular to 4D methods and is extremely challenging: the test views are far away from training views, camera poses are often inaccurate, depths are noisy and training views have issues such as overexposure and autofocus. The dataset consists of 14 casually captured scenes, 7 of which have no ground truth test views and are used for qualitative evaluation only and 7 with test views available. Due to the difficulty of obtaining accurate camera poses for all scenes, some methods choose to quantitatively evaluate on only 5 of the available 7 scenes and discard `space-out' and `wheel'. To our knowledge, this is currently the only widely used benchmark which is appropriate for evaluating our method. As described in DyCheck \citep{gao2022}, other datasets such as Nerfies \citep{park2021b} , HyperNeRF \citep{park2021a} and NSFF \citep{li2021} suffer from teleporting cameras which makes them effectively multi-view. As described in MoSca \citep{lei2024}, the NVIDIA dataset \cite{yoon2020novel} is forward-facing with small-baseline static cameras and is significantly easier than DyCheck, thus our contributions which tackle highly ill-posed settings are less useful. We quantitatively and qualitatively evaluate our method and other state of the art baselines across all 14 scenes.

\begin{table}
\centering
\caption{Quantitative results on dynamic regions of scenes from the DyCheck (iPhone) dataset. Best, second and third results are highlighted in red, orange and yellow respectively.   SoM-5 is full-res with wheel and space-out excluded.}
\begin{tabularx}{\linewidth}{llZZZ}  
    \toprule
     & Method & PSNR-D\,\uparrowaligned & SSIM-D\,\uparrowaligned & LPIPS-D\,\downarrowaligned  \\ 
      \midrule
      \multirow{6}{*}{\STAB{\rotatebox[origin=c]{90}{Half-Res}}} & T-NeRF\,\citep{gao2022} & 13.86 & 0.8546 & 0.3491 \\ 
      & Nerfies\,\citep{park2021b} & 12.89 & 0.8425 & 0.3811\\ 
      & HyperNeRF\,\citep{park2021a} & 13.27 & 0.8484 & 0.3558\\ 
      &Gaussian Marbles\,\citep{stearns2024} & 9.99 & 0.8175 & 0.3926 \\ 
      & SoM\,\citep{wang2024}  & \cellthird14.80  & \cellthird0.8582 & \cellthird0.3008 \\ 
    &MoSca\,\citep{lei2024} & \cellsecond15.63 & \cellsecond0.8755 & \cellsecond0.2904 \\
    &Ours & \cellbest16.46 & \cellbest0.8850 & \cellbest0.2793\\
    \midrule
      \multirow{4}{*}{\STAB{\rotatebox[origin=c]{90}{Full-Res}}} &  Gaussian Marbles\,\citep{stearns2024} &  12.75 & 0.8607 & 0.5058  \\  
      & SoM\,\citep{wang2024}  & \cellthird14.82 & \cellthird0.8709 & \cellsecond0.4347\\ 
    &MoSca\,\citep{lei2024} & \cellsecond15.39 & \cellsecond 0.8821 &  \cellthird0.4413 \\
    &Ours & \cellbest16.32 & \cellbest0.8893 & \cellbest0.3921 \\
    \midrule
      \multirow{4}{*}{\STAB{\rotatebox[origin=c]{90}{SoM-5}}} &  Gaussian Marbles\,\citep{stearns2024} & \cellthird13.66 & \cellthird0.8658 & 0.4919 \\ 
    & SoM\,\citep{wang2024}  & 12.50 & 0.8648 & \cellthird0.4890\\ 
    &MoSca\,\citep{lei2024} & \cellsecond15.83& \cellsecond0.8872& \cellsecond0.4404\\ 
    &Ours & \cellbest16.69 &\cellbest0.8941&\cellbest0.3778 \\
    \bottomrule
\end{tabularx}
\label{tab:results-dynamic}
\end{table}

\subsection{Metrics}
Following previous works \citep{gao2022,lei2024}, we compute PSNR, SSIM and LPIPS on the co-visibility masked regions of the test views, which we denote with an \textit{-m} addendum to each metric. We compute metrics at both half-resolution and full-resolution, and following \citep{wang2024}, we also report results on a subset of 5 scenes which we label SoM-5. 
\subsubsection{Limitations of Metrics and A New Benchmark}
\begin{figure}
    \begin{minipage}{0.55\linewidth}
        \centering
        \captionof{table}{Intersection of co-visibility mask with dynamic regions with respect to co-visibility mask area}
        \begin{tabularx}{0.9\linewidth}{lZ}  
            \toprule
             Scene & Dyn/Co-vis Intersection (\%) \\ 
              \midrule
              apple & 4.42 \\ 
              block & 27.46\\ 
              paper-windmill& 3.58 \\ 
              space-out &  20.63 \\ 
              spin &  19.76 \\ 
              teddy &  81.33 \\ 
              wheel &   24.65\\ 
              mean &  25.97\\ 
            \bottomrule
        \end{tabularx}
        \label{tab:covis}
    \end{minipage}
    \quad
    \begin{minipage}{0.4\linewidth}
        \centering
        \includegraphics[width=0.49\textwidth]{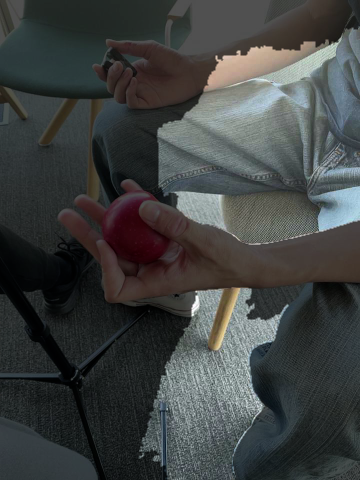}
        \includegraphics[width=0.49\textwidth]{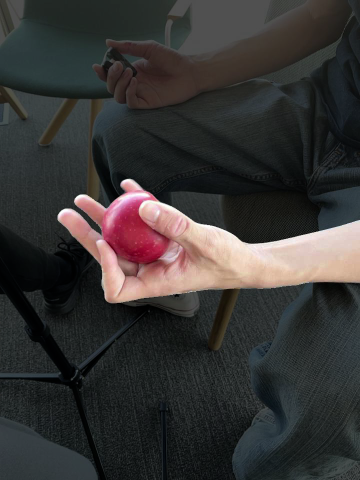}
        \begin{tabularx}{0.99\linewidth}{YY}
        \small{Co-visibility mask} & \small{Dynamic mask}
    \end{tabularx}
        \vspace{-3mm}
        \caption{An example of co-visibility and dynamic mask comparison.}
        \label{fig:covis_dyn_comp}
    \end{minipage}
\end{figure}

We note that the static regions of a scene are often observed from several viewpoints across different time steps in the captured monocular video. This effectively provides multi-view supervision for these regions, and although we are interested in reconstructing the entire scene which includes the static regions, the dynamic regions are arguably the area of most interest and also the most under-observed. In order to better evaluate performance in the dynamic regions of the scene, we compute a set of dynamic masks for each scene using Track Anything \citep{yang2023}. We compute the intersection between the co-visibility masks and the dynamic regions of the scene and present results in Table~\ref{tab:covis}. We find that on average only 26\% of the co-visibility masked pixels correspond to the dynamic region. Some scenes such as apple and paper-windmill have an intersection as low as 4\%. We show an example of this in Figure~\ref{fig:covis_dyn_comp}. The co-visibility masked metrics are heavily weighted towards the static regions of the scene. Although this is useful for evaluating overall reconstruction performance, it underweights the reconstruction performance of methods in the most difficult dynamic regions. We provide a complementary new benchmark for the evaluation of monocular to 4D reconstruction methods, where our computed dynamic masks can be used in place of the commonly used co-visibility masks. We use these masks to compute the PSNR, SSIM and LPIPS, which we denote with a \textit{-D} addendum, for a range of baseline methods in Table~\ref{tab:results-dynamic}.

\begin{figure}
    \centering
    \includegraphics[width=0.16\linewidth]{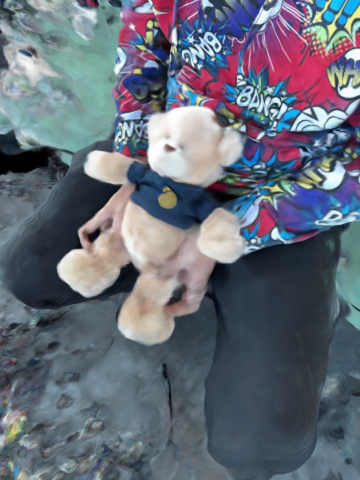}
    \includegraphics[width=0.16\linewidth]{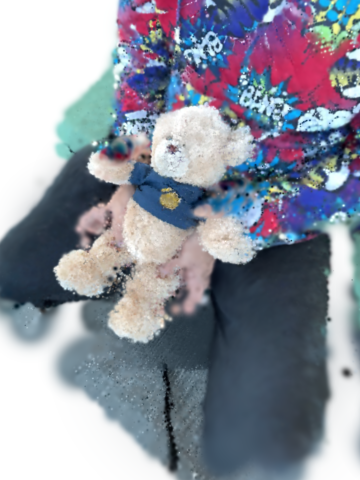}
    \includegraphics[width=0.16\linewidth]{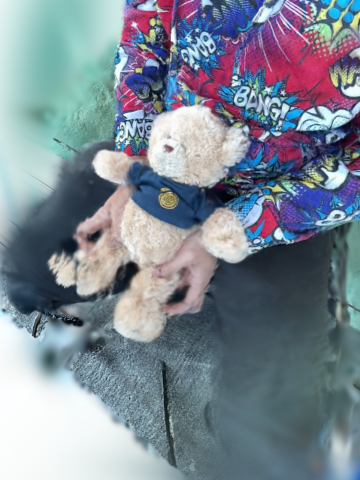}
    \includegraphics[width=0.16\linewidth]{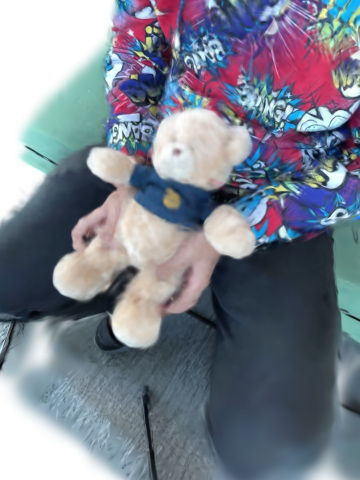}
    \includegraphics[width=0.16\linewidth]{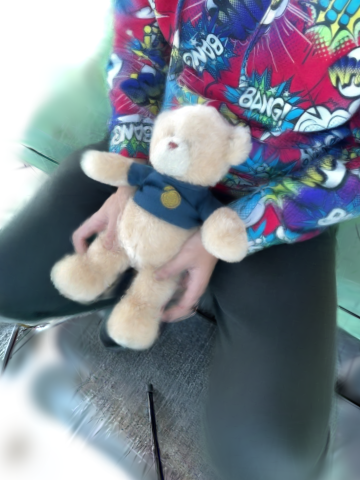}
    \includegraphics[width=0.16\linewidth]{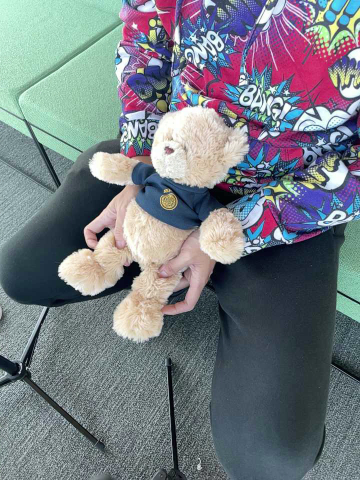}

    \includegraphics[width=0.16\linewidth]{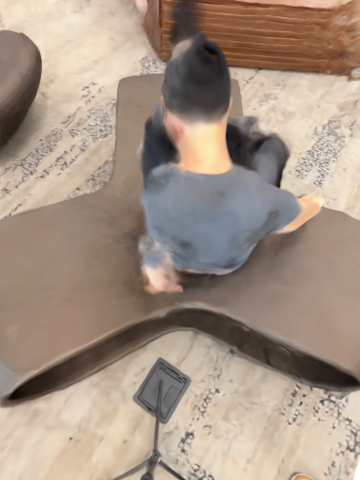}
    \includegraphics[width=0.16\linewidth]{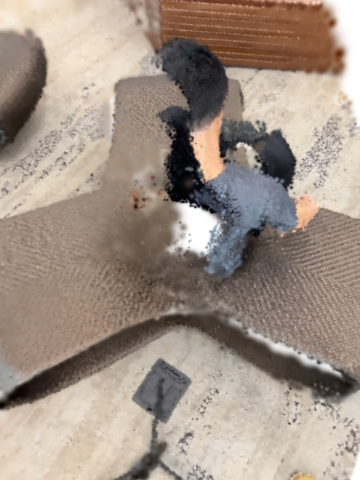}
    \includegraphics[width=0.16\linewidth]{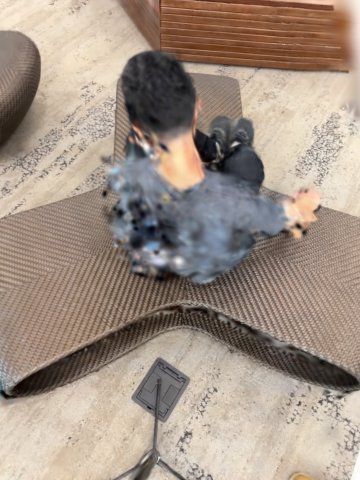}
    \includegraphics[width=0.16\linewidth]{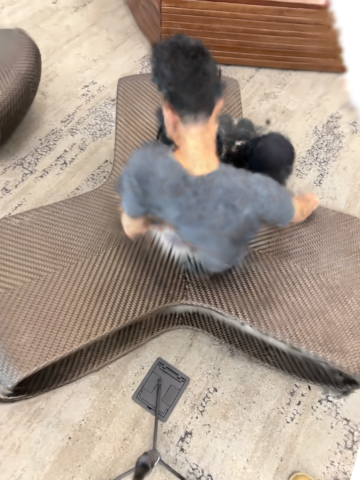}
    \includegraphics[width=0.16\linewidth]{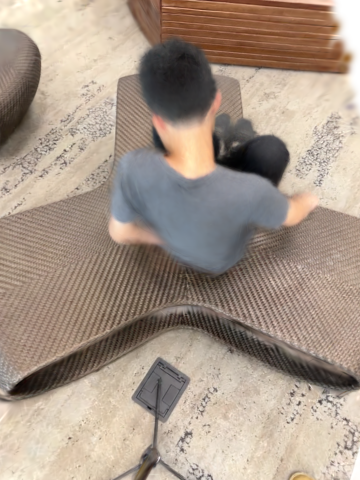}
    \includegraphics[width=0.16\linewidth]{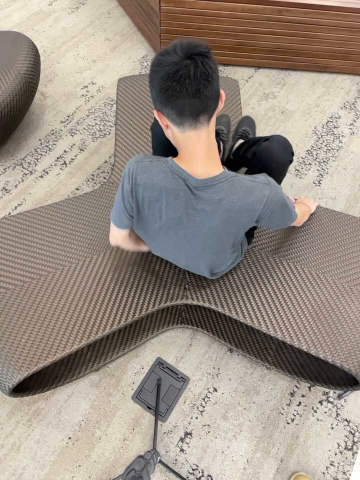}

    \includegraphics[width=0.16\linewidth]{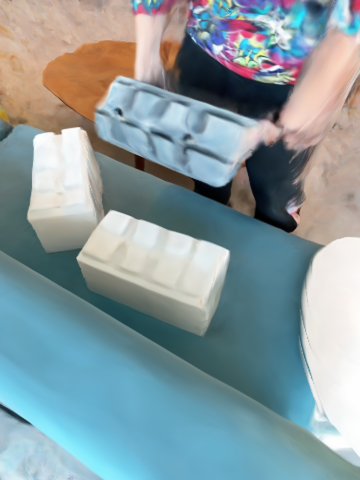}
    \includegraphics[width=0.16\linewidth]{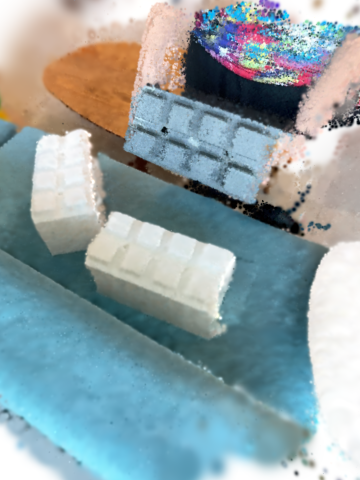}
    \includegraphics[width=0.16\linewidth]{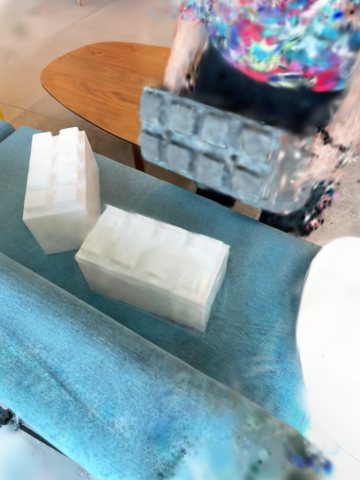}
    \includegraphics[width=0.16\linewidth]{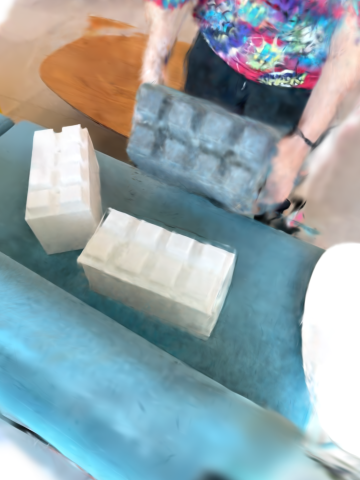}
    \includegraphics[width=0.16\linewidth]{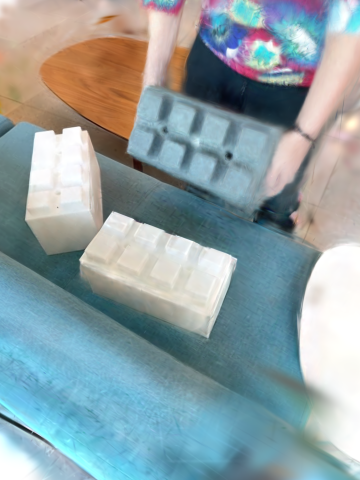}
    \includegraphics[width=0.16\linewidth]{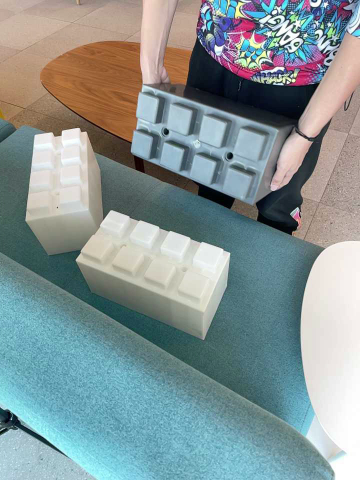}

    \includegraphics[width=0.16\linewidth]{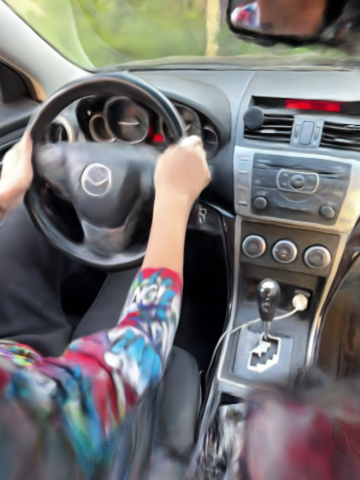}
    \includegraphics[width=0.16\linewidth]{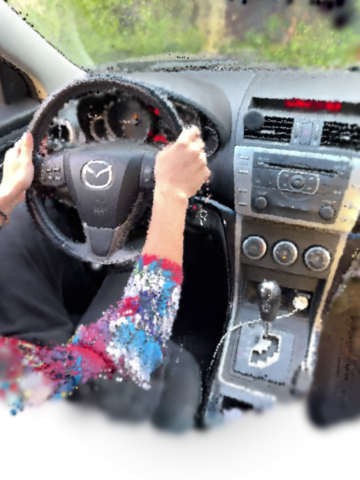}
    \includegraphics[width=0.16\linewidth]{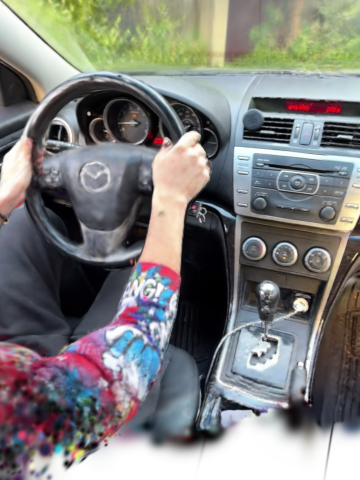}
    \includegraphics[width=0.16\linewidth]{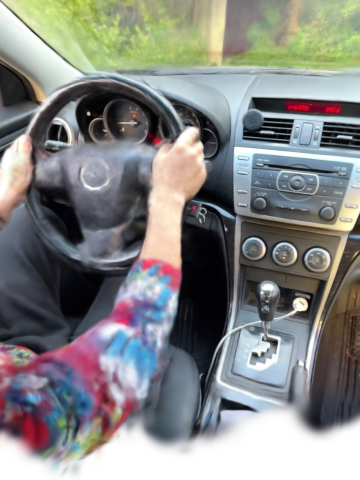}
    \includegraphics[width=0.16\linewidth]{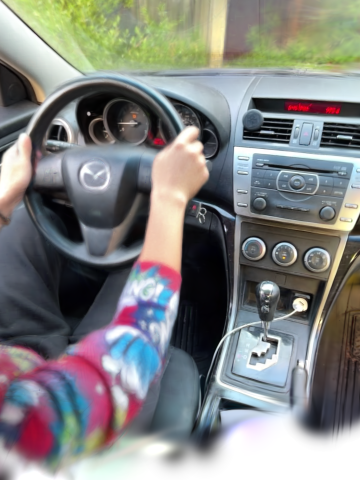}
    \includegraphics[width=0.16\linewidth]{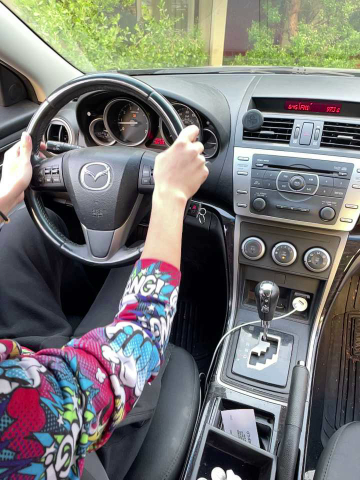}

    \includegraphics[width=0.16\linewidth]{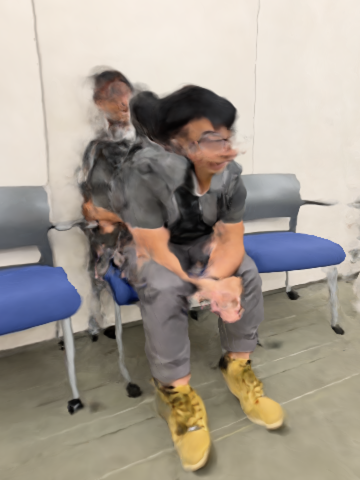}
    \includegraphics[width=0.16\linewidth]{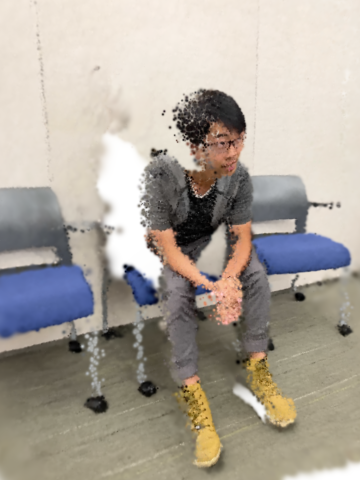}
    \includegraphics[width=0.16\linewidth]{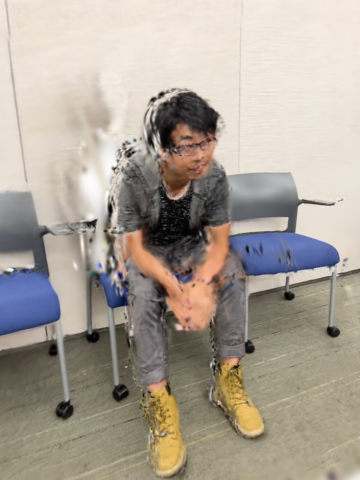}
    \includegraphics[width=0.16\linewidth]{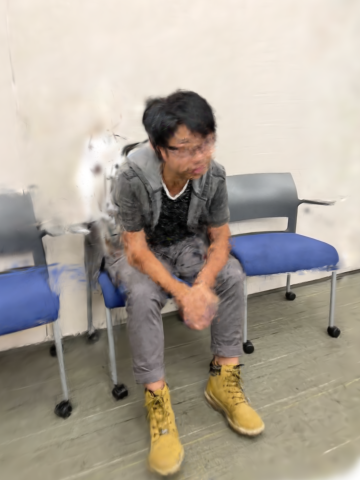}
    \includegraphics[width=0.16\linewidth]{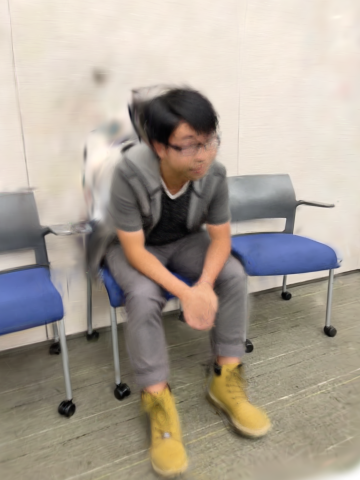}
    \includegraphics[width=0.16\linewidth]{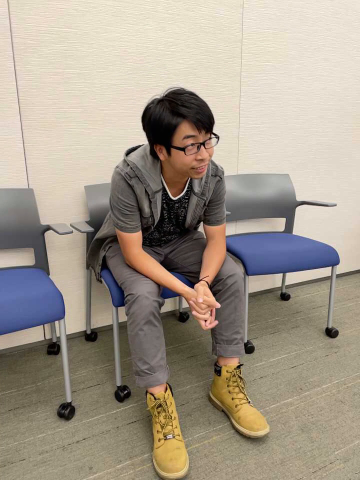}

    \includegraphics[width=0.16\linewidth]{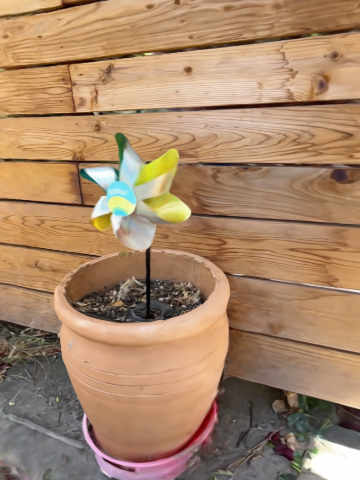}
    \includegraphics[width=0.16\linewidth]{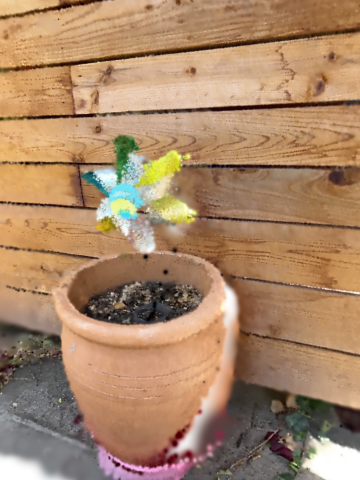}
    \includegraphics[width=0.16\linewidth]{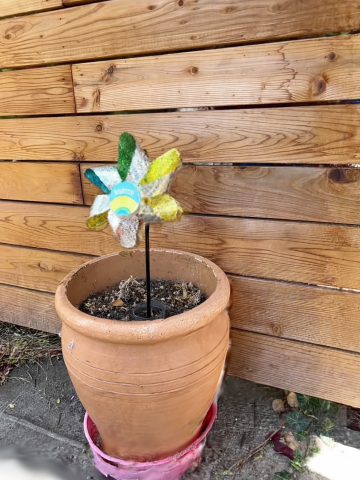}
    \includegraphics[width=0.16\linewidth]{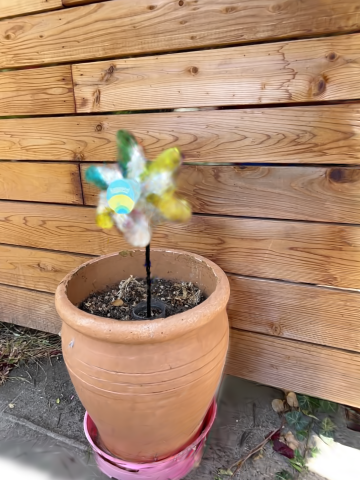}
    \includegraphics[width=0.16\linewidth]{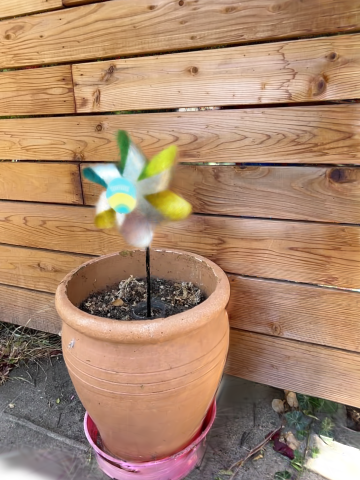}
    \includegraphics[width=0.16\linewidth]{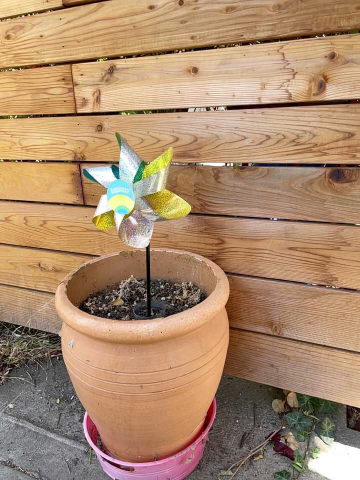}

    \includegraphics[width=0.16\linewidth]{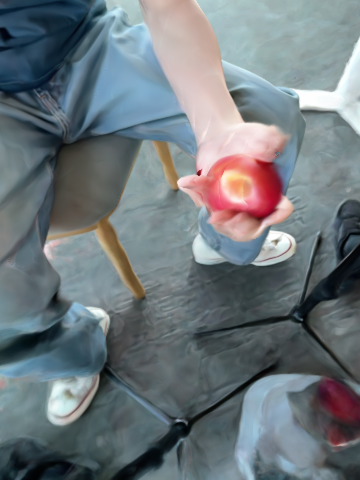}
    \includegraphics[width=0.16\linewidth]{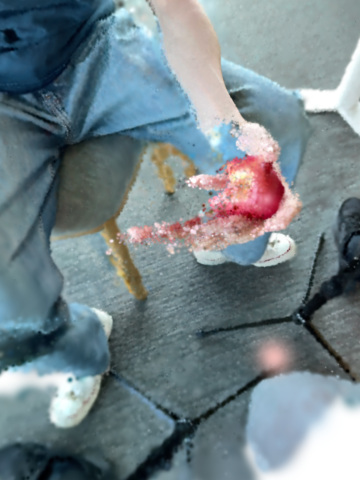}
    \includegraphics[width=0.16\linewidth]{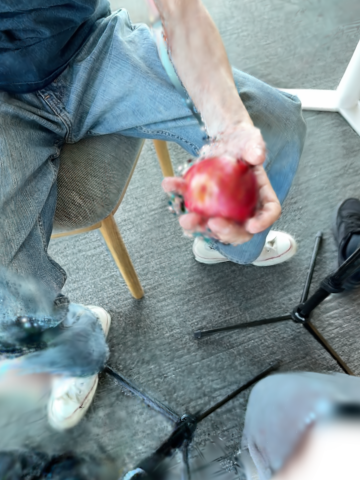}
    \includegraphics[width=0.16\linewidth]{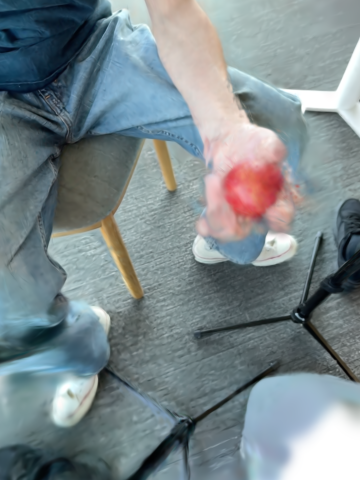}
    \includegraphics[width=0.16\linewidth]{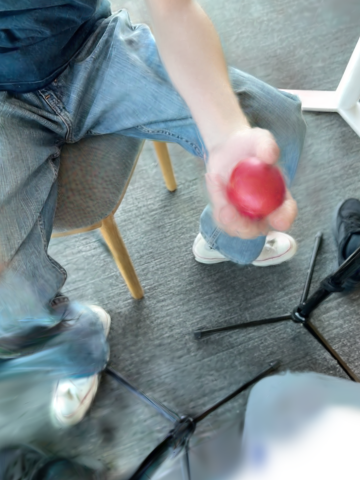}
    \includegraphics[width=0.16\linewidth]{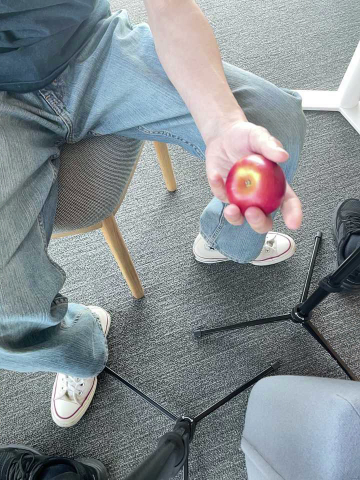}
    \small
    \begin{tabularx}{0.99\linewidth}{YYYYYY} 
        T-NeRF\,\citep{gao2022} & Gaussian Marbles\,\citep{stearns2024} & Shape Of Motion\,\citep{wang2024} & MoSca\,\cite{lei2024} & Ours & GT
    \end{tabularx}
    \caption{Qualitative evaluation of our method against benchmark methods on the DyCheck test set.}
    \label{fig:quali-1}
\end{figure}

\subsection{Evaluation}

\paragraph{Baselines}
We compare against a wide range of baselines, including a number of recent state-of-the-art methods such as MoSca \citep{lei2024}, CAT4D \citep{wu2024b}, Shape Of Motion \citep{wang2024}, Dynamic Gaussian Marbles \citep{stearns2024} and 4DGS \citep{wu2024}, which are based upon Gaussian Splatting \cite{kerbl2023}. We also compare against NeRF-based approaches T-NeRF \citep{gao2022}, Nerfies \citep{park2021b}, HyperNeRF \citep{park2021a}, DyBluRF \citep{bui2023} and RoDynRF \citep{liu2023}, neural point clouds approaches DynPoint \citep{zhou2023} and D-NPC \citep{kappel2025}, generalized pre-trained transformer PGDVS \citep{zhao2024}, neural scene flow NSFF \citep{li2021} and volumetric image-based rendering DynIBaR \citep{li2023}.

\paragraph{Quantitative and Qualitative Evaluation}
We present quantitative results of our method in Tables \ref{tab:results-covis} and \ref{tab:results-dynamic}. Our method outperforms all state-of-the-art baselines in PSNR and SSIM and all but one in LPIPS, across all settings and resolutions. We typically improve PSNR by a large margin, achieving a minimum of 1dB improvement over all methods, except for MoSca where we average 0.94dB and 0.56dB higher in dynamic and co-visibility masked regions respectively. This indicates our method particularly improves dynamic region reconstruction. We note that CAT4D achieves a lower LPIPS score than our method, but the improved perceptual quality comes at the cost of reduced spatio-temporal consistency, which is reflected in the PSNR and SSIM scores, and also clearly shown in our supplementary video. We present a qualitative evaluation in Figure \ref{fig:quali-1} where \name~demonstrates consistently superior visual quality and geometric consistency when compared to the best existing approaches. Although 2D image comparisons are indicative of performance, we encourage viewing our supplementary video results to appreciate the improvement in spatio-temporal consistency and visual quality over baselines.

\begin{figure}
    \centering
    \includegraphics[width=0.16\linewidth]{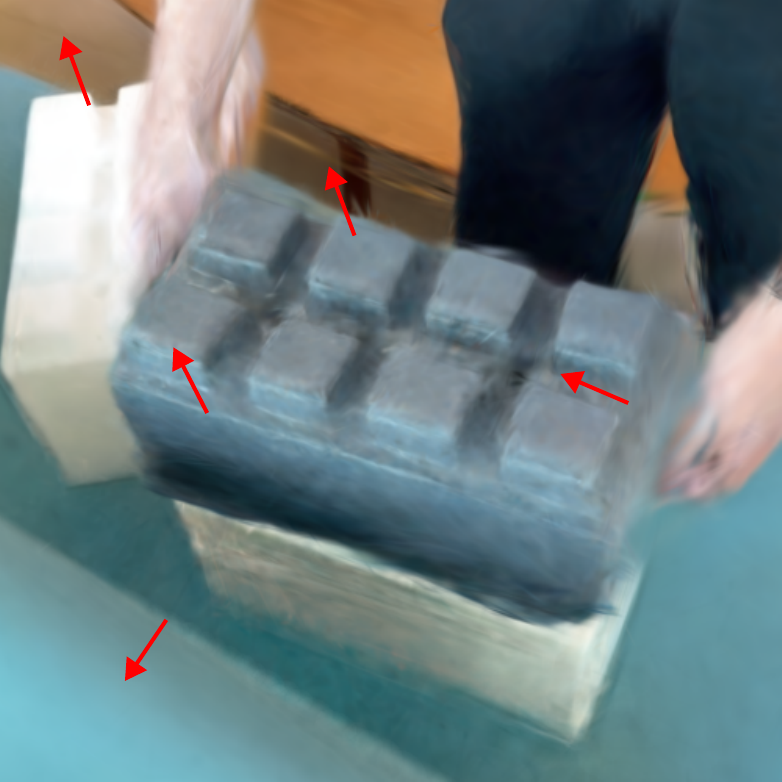}
    \includegraphics[width=0.16\linewidth]{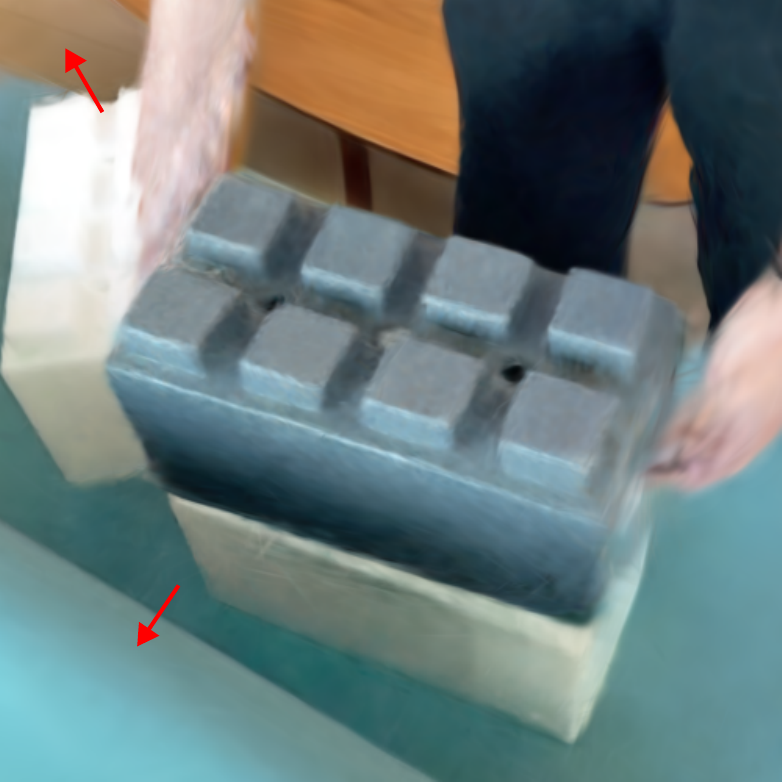}
    \includegraphics[width=0.16\linewidth]{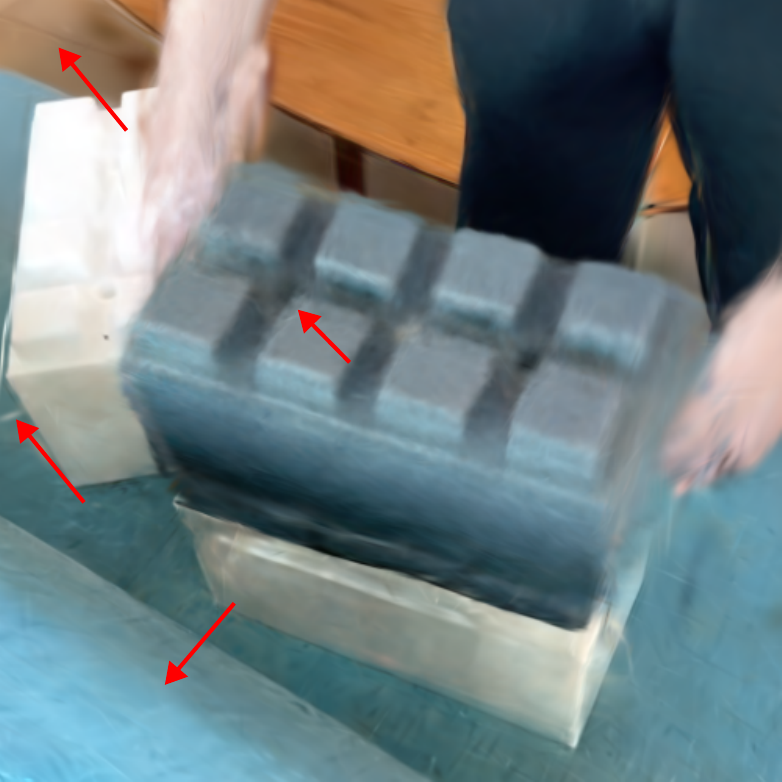}
    \includegraphics[width=0.16\linewidth]{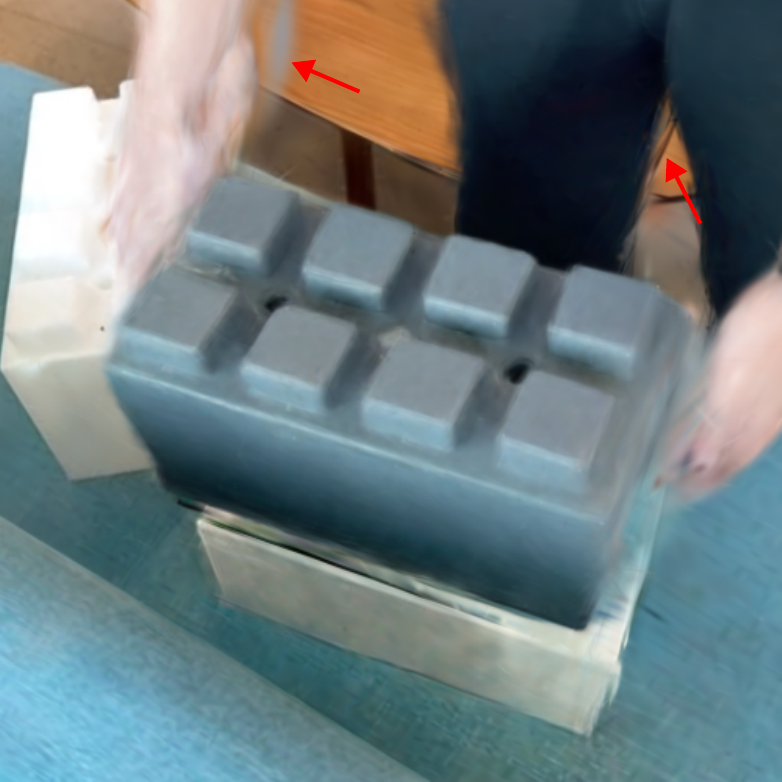}
    \includegraphics[width=0.16\linewidth]{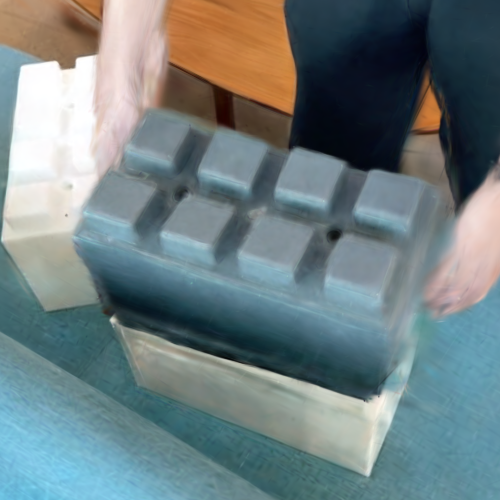}
    \includegraphics[width=0.16\linewidth]{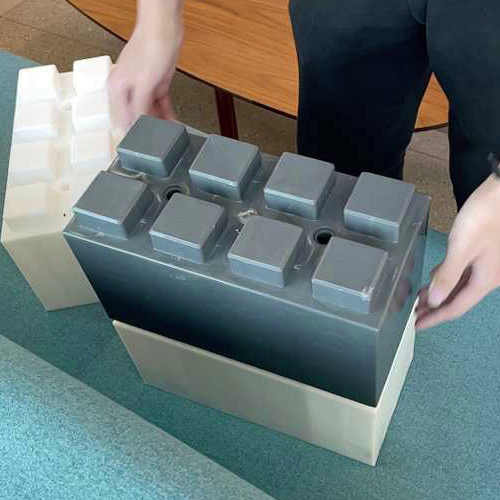}
    \footnotesize
    \setlength{\tabcolsep}{1pt}
    \begin{tabularx}{\linewidth}{YYYYYY} 
        W/o SO/DR/TGS & W/o DR &  W/o SO &  W/o TGS & Ours & GT
    \end{tabularx}
    \caption{Qualitative evaluation of our ablation study with settings corresponding to Tab.~\ref{tab:ablation}.}
    \label{fig:ablation}
\end{figure}

\paragraph{Ablations} We quantitatively evaluate each of our contributions in an ablation study presented in Table~\ref{tab:ablation}. The bottom row w/o SO + DR + TGS shows a naive approach of using the diffused novel views directly as supervision for our monocular baseline without diffusion-aware reconstruction. Due to the spatio-temporal inconsistencies of the diffused outputs, this leads to a poor quality reconstruction, as shown in Figure~\ref{fig:ablation}. We show that removing dynamic reconstruction leads to blurry reconstruction in static regions, while removing sampled camera optimization leads to geometric inconsistencies. We also show that using our tracking based Gaussian classification reduces floaters.

\begin{table}
\centering
\caption{Quantitative results of an ablation study of the components of \name{}.}
\begin{tabularx}{\linewidth}{lZZZ}  
    \toprule
     Method &  PSNR-m & SSIM-m & LPIPS-m \\ 
      \midrule
      Ours & \cellbest19.00 & \cellbest0.6672& \cellbest0.3623 \\ 
      W/o Tracking Based Gaussian Classification (TGS)& \cellthird18.88 &\cellsecond0.6651 & \cellsecond0.3693 \\ 
      W/o Sampled Camera Optimisation (SO) &18.39  &\cellthird0.6514 & \cellthird0.4040 \\ 
      W/o Dynamic Reconstruction (DR)&  \cellsecond18.93&0.6274 & 0.4497 \\ 
      W/o SO + DR + TGS & 18.46 &0.6075 & 0.4656 \\ 
    \bottomrule
\end{tabularx}
\label{tab:ablation}
\end{table}

\section{Conclusion} \label{sec:conclusions}

We present ViDAR, a novel method for 4D reconstruction of scenes from monocular inputs. ViDAR leverages video diffusion models by conditioning on scene-specific features to recover fine-grained appearance details of novel viewpoints. ViDAR overcomes the spatio-temporal inconsistency of diffusion-based supervision via a diffusion-aware loss function and a camera pose optimisation strategy. We show that ViDAR outperforms all state-of-the-art baselines on the challening DyCheck dataset, and we present a new benchmark to evaluate performance in dynamic regions.

\textbf{Limitations:}
\name{} limits the scope of diffusion to enhancing rendered images, which are limited by the initial accuracy of the 4D reconstruction, thus, cannot repair major geometrical artefacts.

{
    \small
    \bibliographystyle{abbrvnat}
    \bibliography{references}

\begin{thebibliography}{64}
\providecommand{\natexlab}[1]{#1}
\providecommand{\url}[1]{\texttt{#1}}
\expandafter\ifx\csname urlstyle\endcsname\relax
  \providecommand{\doi}[1]{doi: #1}\else
  \providecommand{\doi}{doi: \begingroup \urlstyle{rm}\Url}\fi

\bibitem[Bui et~al.(2023)Bui, Park, Oh, and Kim]{bui2023}
M.-Q.~V. Bui, J.~Park, J.~Oh, and M.~Kim.
\newblock {DyBluRF: Dynamic Deblurring Neural Radiance Fields for Blurry Monocular Video}.
\newblock \emph{arXiv preprint arXiv:2312.13528}, 2023.

\bibitem[Cao and CV(2023)]{cao2023}
A.~Cao and J.~CV.
\newblock {HexPlane: A Fast Representation for Dynamic Scenes}.
\newblock In \emph{Computer Vision and Pattern Recognition Conference (CVPR)}, 2023.

\bibitem[Fridovich-Keil et~al.(2023)Fridovich-Keil, Meanti, Warburg, Recht, and Kanazawa]{fridovich2023}
S.~Fridovich-Keil, G.~Meanti, F.~R. Warburg, B.~Recht, and A.~Kanazawa.
\newblock {K-Planes: Explicit Radiance Fields in Space, Time, and Appearance}.
\newblock In \emph{Computer Vision and Pattern Recognition Conference (CVPR)}, 2023.

\bibitem[Gao et~al.(2022)Gao, Li, Tulsiani, Russell, and Kanazawa]{gao2022}
H.~Gao, R.~Li, S.~Tulsiani, B.~Russell, and A.~Kanazawa.
\newblock {Monocular Dynamic View Synthesis: A Reality Check}.
\newblock In \emph{Conference on Neural Information Processing Systems}, 2022.

\bibitem[Guizilini et~al.(2025)Guizilini, Irshad, Chen, Shakhnarovich, and Ambrus]{guizilini2025}
V.~Guizilini, M.~Z. Irshad, D.~Chen, G.~Shakhnarovich, and R.~Ambrus.
\newblock {Zero-Shot Novel View and Depth Synthesis with Multi-View Geometric Diffusion}.
\newblock In \emph{Computer Vision and Pattern Recognition Conference (CVPR)}, 2025.

\bibitem[Huang et~al.(2023)Huang, Sun, Yang, Lyu, Cao, and Qi]{huang2024}
Y.-H. Huang, Y.-T. Sun, Z.~Yang, X.~Lyu, Y.-P. Cao, and X.~Qi.
\newblock {SC-GS: Sparse-Controlled Gaussian Splatting for Editable Dynamic Scenes}.
\newblock In \emph{Computer Vision and Pattern Recognition Conference (CVPR)}, 2023.

\bibitem[Kappel et~al.(2025)Kappel, Hahlbohm, Scholz, Castillo, Theobalt, Eisemann, Golyanik, and Magnor]{kappel2025}
M.~Kappel, F.~Hahlbohm, T.~Scholz, S.~Castillo, C.~Theobalt, M.~Eisemann, V.~Golyanik, and M.~Magnor.
\newblock D-{NPC}: Dynamic neural point clouds for non-rigid view synthesis from monocular video.
\newblock \emph{Proceedings of the Eurographics Conference (EG)}, 44, 2025.

\bibitem[Karras et~al.(2022)Karras, Aittala, Aila, and Laine]{karras2022}
T.~Karras, M.~Aittala, T.~Aila, and S.~Laine.
\newblock Elucidating the design space of diffusion-based generative models.
\newblock In \emph{Conference on Neural Information Processing Systems}, 2022.

\bibitem[Kerbl et~al.(2023)Kerbl, Kopanas, Leimk{\"u}hler, and Drettakis]{kerbl2023}
B.~Kerbl, G.~Kopanas, T.~Leimk{\"u}hler, and G.~Drettakis.
\newblock {3D Gaussian Splatting for Real-Time Radiance Field Rendering}.
\newblock \emph{ACM Transactions on Graphics}, 42\penalty0 (4), July 2023.

\bibitem[Lee et~al.(2024)Lee, Won, Jung, Bae, and Jeon]{lee2024}
J.~Lee, C.~Won, H.~Jung, I.~Bae, and H.-G. Jeon.
\newblock {Fully Explicit Dynamic Guassian Splatting}.
\newblock In \emph{Proceedings of the Neural Information Processing Systems}, 2024.

\bibitem[Lei et~al.(2025)Lei, Weng, Harley, Guibas, and Daniilidis]{lei2024}
J.~Lei, Y.~Weng, A.~Harley, L.~Guibas, and K.~Daniilidis.
\newblock {MoSca}: Dynamic gaussian fusion from casual videos via {4D} motion scaffolds.
\newblock \emph{Computer Vision and Pattern Recognition Conference (CVPR)}, 2025.

\bibitem[Li et~al.(2024{\natexlab{a}})Li, Shi, Zhang, Wu, Liao, Wang, Lee, and Zhou]{li2024a}
H.~Li, H.~Shi, W.~Zhang, W.~Wu, Y.~Liao, L.~Wang, L.-h. Lee, and P.~Y. Zhou.
\newblock {Dreamscene: 3d gaussian-based text-to-3d scene generation via formation pattern sampling}.
\newblock In \emph{European Conference on Computer Vision (ECCV)}, 2024{\natexlab{a}}.

\bibitem[Li et~al.(2022)Li, Shen, Wang, Shen, and Tan]{li2022b}
L.~Li, Z.~Shen, Z.~Wang, L.~Shen, and P.~Tan.
\newblock {Streaming Radiance Fields for 3D Video Synthesis}.
\newblock In \emph{Conference on Neural Information Processing Systems}, 2022.

\bibitem[Li et~al.(2021)Li, Niklaus, Snavely, and Wang]{li2021}
Z.~Li, S.~Niklaus, N.~Snavely, and O.~Wang.
\newblock {Neural Scene Flow Fields for Space-Time View Synthesis of Dynamic Scenes}.
\newblock In \emph{Computer Vision and Pattern Recognition Conference (CVPR)}, 2021.

\bibitem[Li et~al.(2023)Li, Wang, Cole, Tucker, and Snavely]{li2023}
Z.~Li, Q.~Wang, F.~Cole, R.~Tucker, and N.~Snavely.
\newblock {DynIBaR: Neural Dynamic Image-Based Rendering}.
\newblock In \emph{Computer Vision and Pattern Recognition Conference (CVPR)}, 2023.

\bibitem[Li et~al.(2024{\natexlab{b}})Li, Chen, Li, and Xu]{li2024}
Z.~Li, Z.~Chen, Z.~Li, and Y.~Xu.
\newblock {Spacetime Gaussian Feature Splatting for Real-Time Dynamic View Synthesis}.
\newblock In \emph{Computer Vision and Pattern Recognition Conference (CVPR)}, 2024{\natexlab{b}}.

\bibitem[Liang et~al.(2024{\natexlab{a}})Liang, Ren, Mirzaei, Torralba, Liu, Gilitschenski, Fidler, Oztireli, Ling, Gojcic, and Huang]{liang2024}
H.~Liang, J.~Ren, A.~Mirzaei, A.~Torralba, Z.~Liu, I.~Gilitschenski, S.~Fidler, C.~Oztireli, H.~Ling, Z.~Gojcic, and J.~Huang.
\newblock {Feed-Forward Bullet-Time Reconstruction of Dynamic Scenes from Monocular Videos}.
\newblock \emph{arXiv preprint arXiv:2412.03526}, 2024{\natexlab{a}}.

\bibitem[Liang et~al.(2024{\natexlab{b}})Liang, Yang, Lin, Li, Xu, and Chen]{liang2024b}
Y.~Liang, X.~Yang, J.~Lin, H.~Li, X.~Xu, and Y.~Chen.
\newblock {Luciddreamer: Towards high-fidelity text-to-3d generation via interval score matching}.
\newblock In \emph{Computer Vision and Pattern Recognition Conference (CVPR)}, 2024{\natexlab{b}}.

\bibitem[Lin et~al.(2025)Lin, Pan, Yang, Li, and Mu]{lin2025}
C.~Lin, P.~Pan, B.~Yang, Z.~Li, and Y.~Mu.
\newblock {DiffSplat: Repurposing Image Diffusion Models for Scalable 3D Gaussian Splat Generation}.
\newblock In \emph{International Conference on Learning Representations (ICLR)}, 2025.

\bibitem[Lin et~al.(2024)Lin, Dai, Zhu, and Yao]{lin2024}
Y.~Lin, Z.~Dai, S.~Zhu, and Y.~Yao.
\newblock {Gaussian-Flow: 4D Reconstruction with Dynamic 3D Gaussian Particle}.
\newblock In \emph{Computer Vision and Pattern Recognition Conference (CVPR)}, 2024.

\bibitem[Liu et~al.(2024{\natexlab{a}})Liu, Shi, Chen, Zhang, Xu, Wei, Chen, Zeng, Gu, and Su]{liu2024a}
M.~Liu, R.~Shi, L.~Chen, Z.~Zhang, C.~Xu, X.~Wei, H.~Chen, C.~Zeng, J.~Gu, and H.~Su.
\newblock {One-2-3-45++: Fast Single Image to 3D Objects with Consistent Multi-View Generation and 3D Diffusion}.
\newblock In \emph{Computer Vision and Pattern Recognition Conference (CVPR)}, 2024{\natexlab{a}}.

\bibitem[Liu et~al.(2024{\natexlab{b}})Liu, Zeng, Wei, Shi, Chen, Xu, Zhang, Wang, Zhang, Liu, Wu, and Su]{liu2024}
M.~Liu, C.~Zeng, X.~Wei, R.~Shi, L.~Chen, C.~Xu, M.~Zhang, Z.~Wang, X.~Zhang, I.~Liu, H.~Wu, and H.~Su.
\newblock {MeshFormer: High-Quality Mesh Generation with 3D-Guided Reconstruction Model}.
\newblock In \emph{Conference on Neural Information Processing Systems}, 2024{\natexlab{b}}.

\bibitem[Liu et~al.(2025)Liu, Liu, Wang, Lyu, Wang, Wang, and Hou]{liu2025}
Q.~Liu, Y.~Liu, J.~Wang, X.~Lyu, P.~Wang, W.~Wang, and J.~Hou.
\newblock {MoDGS: Dynamic Gaussian Splatting from Casually-captured Monocular Videos with Depth Priors}.
\newblock In \emph{International Conference on Learning Representations (ICLR)}, 2025.

\bibitem[Liu et~al.(2023)Liu, Gao, Meuleman, Tseng, Saraf, Kim, Chuang, Kopf, and Huang]{liu2023}
Y.-L. Liu, C.~Gao, A.~Meuleman, H.-Y. Tseng, A.~Saraf, C.~Kim, Y.-Y. Chuang, J.~Kopf, and J.-B. Huang.
\newblock Robust dynamic radiance fields.
\newblock In \emph{Computer Vision and Pattern Recognition Conference (CVPR)}, 2023.

\bibitem[Luiten et~al.(2024)Luiten, Kopanas, Leibe, and Ramanan]{luiten2023}
J.~Luiten, G.~Kopanas, B.~Leibe, and D.~Ramanan.
\newblock {Dynamic 3D Gaussians: Tracking by Persistent Dynamic View Synthesis}.
\newblock In \emph{International Conference on 3D Vision (3DV)}, 2024.

\bibitem[Miao et~al.(2024)Miao, Bai, Duan, Wan, Huang, Long, and Zheng]{miao2024}
X.~Miao, Y.~Bai, H.~Duan, F.~Wan, Y.~Huang, Y.~Long, and Y.~Zheng.
\newblock {CTNeRF: Cross-time Transformer for dynamic neural radiance field from monocular video}.
\newblock \emph{Pattern Recognition}, 156:\penalty0 110729, 2024.

\bibitem[Mildenhall et~al.(2020)Mildenhall, Srinivasan, Tancik, Barron, Ramamoorthi, and Ng]{mildenhall2020}
B.~Mildenhall, P.~P. Srinivasan, M.~Tancik, J.~T. Barron, R.~Ramamoorthi, and R.~Ng.
\newblock {NeRF}: Representing scenes as neural radiance fields for view synthesis.
\newblock In \emph{European Conference on Computer Vision (ECCV)}, 2020.

\bibitem[Park et~al.(2025)Park, Bui, Bello, Moon, Oh, and Kim]{park2025}
J.~Park, M.-Q.~V. Bui, J.~L.~G. Bello, J.~Moon, J.~Oh, and M.~Kim.
\newblock {SplineGS: Robust Motion-Adaptive Spline for Real-Time Dynamic 3D Gaussians from Monocular Video}.
\newblock In \emph{Computer Vision and Pattern Recognition Conference (CVPR)}, 2025.

\bibitem[Park et~al.(2021{\natexlab{a}})Park, Sinha, Barron, Bouaziz, Goldman, Seitz, and Martin-Brualla]{park2021b}
K.~Park, U.~Sinha, J.~T. Barron, S.~Bouaziz, D.~B. Goldman, S.~M. Seitz, and R.~Martin-Brualla.
\newblock {Nerfies: Deformable Neural Radiance Fields}.
\newblock \emph{International Conference on Computer Vision (ICCV)}, 2021{\natexlab{a}}.

\bibitem[Park et~al.(2021{\natexlab{b}})Park, Sinha, Hedman, Barron, Bouaziz, Goldman, Martin-Brualla, and Seitz]{park2021a}
K.~Park, U.~Sinha, P.~Hedman, J.~T. Barron, S.~Bouaziz, D.~B. Goldman, R.~Martin-Brualla, and S.~M. Seitz.
\newblock {HyperNeRF: a higher-dimensional representation for topologically varying neural radiance fields}.
\newblock \emph{ACM Transactions on Graphics}, 40\penalty0 (6), 2021{\natexlab{b}}.

\bibitem[Podell et~al.(2024)Podell, English, Lacey, Blattmann, Dockhorn, Müller, Penna, and Rombach]{podell2024}
D.~Podell, Z.~English, K.~Lacey, A.~Blattmann, T.~Dockhorn, J.~Müller, J.~Penna, and R.~Rombach.
\newblock {SDXL: Improving Latent Diffusion Models for High-Resolution Image Synthesis}.
\newblock In \emph{International Conference on Learning Representations (ICLR)}, 2024.

\bibitem[Poole et~al.(2023)Poole, Jain, Barron, and Mildenhall]{poole2023}
B.~Poole, A.~Jain, J.~T. Barron, and B.~Mildenhall.
\newblock {DreamFusion: Text-to-3D using 2D Diffusion}.
\newblock In \emph{International Conference on Learning Representations (ICLR)}, 2023.

\bibitem[Pumarola et~al.(2021)Pumarola, Corona, Pons-Moll, and Moreno-Noguer]{pumarola2020}
A.~Pumarola, E.~Corona, G.~Pons-Moll, and F.~Moreno-Noguer.
\newblock {D-NeRF: Neural Radiance Fields for Dynamic Scenes}.
\newblock In \emph{Computer Vision and Pattern Recognition Conference (CVPR)}, 2021.

\bibitem[Rombach et~al.(2021)Rombach, Blattmann, Lorenz, Esser, and Ommer]{rombach2021}
R.~Rombach, A.~Blattmann, D.~Lorenz, P.~Esser, and B.~Ommer.
\newblock High-resolution image synthesis with latent diffusion models.
\newblock In \emph{Computer Vision and Pattern Recognition Conference (CVPR)}, 2021.

\bibitem[Ruiz et~al.(2023)Ruiz, Li, Jampani, Pritch, Rubinstein, and Aberman]{ruiz2023}
N.~Ruiz, Y.~Li, V.~Jampani, Y.~Pritch, M.~Rubinstein, and K.~Aberman.
\newblock {DreamBooth: Fine Tuning Text-to-image Diffusion Models for Subject-Driven Generation}.
\newblock In \emph{Computer Vision and Pattern Recognition Conference (CVPR)}, 2023.

\bibitem[Shao et~al.(2023)Shao, Zheng, Tu, Liu, Zhang, and Liu]{shao2023}
R.~Shao, Z.~Zheng, H.~Tu, B.~Liu, H.~Zhang, and Y.~Liu.
\newblock {Tensor4D: Efficient Neural 4D Decomposition for High-fidelity Dynamic Reconstruction and Rendering}.
\newblock In \emph{Computer Vision and Pattern Recognition Conference (CVPR)}, 2023.

\bibitem[Shaw et~al.(2024)Shaw, Nazarczuk, Song, Moreau, Catley-Chandar, Dhamo, and Pérez-Pellitero]{shaw2024}
R.~Shaw, M.~Nazarczuk, J.~Song, A.~Moreau, S.~Catley-Chandar, H.~Dhamo, and E.~Pérez-Pellitero.
\newblock {SWinGS: Sliding Windows for Dynamic 3D Gaussian Splatting}.
\newblock In \emph{European Conference on Computer Vision (ECCV)}, 2024.

\bibitem[Shriram et~al.(2025)Shriram, Trevithick, Liu, and Ramamoorthi]{shriram2025}
J.~Shriram, A.~Trevithick, L.~Liu, and R.~Ramamoorthi.
\newblock {RealmDreamer: Text-Driven 3D Scene Generation with Inpainting and Depth Diffusion}.
\newblock In \emph{International Conference on 3D Vision (3DV)}, 2025.

\bibitem[Simonyan and Zisserman(2015)]{simonyan15}
K.~Simonyan and A.~Zisserman.
\newblock Very deep convolutional networks for large-scale image recognition.
\newblock In \emph{International Conference on Learning Representations}, 2015.

\bibitem[Stearns et~al.(2024)Stearns, Harley, Uy, Dubost, Tombari, Wetzstein, and Guibas]{stearns2024}
C.~Stearns, A.~W. Harley, M.~Uy, F.~Dubost, F.~Tombari, G.~Wetzstein, and L.~Guibas.
\newblock {Dynamic Gaussian Marbles for Novel View Synthesis of Casual Monocular Videos}.
\newblock In \emph{SIGGRAPH Asia}, 2024.

\bibitem[Tang et~al.(2024)Tang, Chen, Chen, Wang, Zeng, and Liu]{tang2024}
J.~Tang, Z.~Chen, X.~Chen, T.~Wang, G.~Zeng, and Z.~Liu.
\newblock {LGM: Large Multi-View Gaussian Model for High-Resolution 3D Content Creation}.
\newblock In \emph{European Conference on Computer Vision (ECCV)}, 2024.

\bibitem[Wang et~al.(2024{\natexlab{a}})Wang, Zhuang, Siarohin, Cao, Qian, Lee, and Tulyakov]{wang2024b}
C.~Wang, P.~Zhuang, A.~Siarohin, J.~Cao, G.~Qian, H.-Y. Lee, and S.~Tulyakov.
\newblock {Diffusion Priors for Dynamic View Synthesis from Monocular Videos}.
\newblock \emph{arXiv preprint arXiv:2401.05583}, 2024{\natexlab{a}}.

\bibitem[Wang et~al.(2023)Wang, Tan, Li, Tian, and Liu]{wang2023}
F.~Wang, S.~Tan, X.~Li, Z.~Tian, and H.~Liu.
\newblock {Mixed Neural Voxels for Fast Multi-view Video Synthesis}.
\newblock In \emph{International Conference on Computer Vision (ICCV)}, 2023.

\bibitem[Wang et~al.(2024{\natexlab{b}})Wang, Ye, Gao, Zeng, Austin, Li, and Kanazawa]{wang2024}
Q.~Wang, V.~Ye, H.~Gao, W.~Zeng, J.~Austin, Z.~Li, and A.~Kanazawa.
\newblock {Shape of Motion: 4D Reconstruction from a Single Video}.
\newblock In \emph{arXiv preprint arXiv:2407.13764}, 2024{\natexlab{b}}.

\bibitem[Wang et~al.(2004)Wang, Bovik, Sheikh, and Simoncelli]{wang2004}
Z.~Wang, A.~Bovik, H.~Sheikh, and E.~Simoncelli.
\newblock {Image quality assessment: from error visibility to structural similarity}.
\newblock \emph{IEEE TIP}, 13\penalty0 (4), 2004.

\bibitem[Wimmer et~al.(2025)Wimmer, Oechsle, Niemeyer, and Tombari]{wimmer2025}
T.~Wimmer, M.~Oechsle, M.~Niemeyer, and F.~Tombari.
\newblock {Gaussians-to-Life: Text-Driven Animation of 3D Gaussian Splatting Scenes}.
\newblock In \emph{International Conference on 3D Vision (3DV)}, 2025.

\bibitem[Wu et~al.(2024{\natexlab{a}})Wu, Yi, Fang, Xie, Zhang, Wei, Liu, Tian, and Wang]{wu2024}
G.~Wu, T.~Yi, J.~Fang, L.~Xie, X.~Zhang, W.~Wei, W.~Liu, Q.~Tian, and X.~Wang.
\newblock {4D Gaussian Splatting for Real-Time Dynamic Scene Rendering}.
\newblock In \emph{Computer Vision and Pattern Recognition Conference (CVPR)}, 2024{\natexlab{a}}.

\bibitem[Wu et~al.(2024{\natexlab{b}})Wu, Gao, Poole, Trevithick, Zheng, Barron, and Holynski]{wu2024b}
R.~Wu, R.~Gao, B.~Poole, A.~Trevithick, C.~Zheng, J.~T. Barron, and A.~Holynski.
\newblock {CAT4D: Create Anything in 4D with Multi-View Video Diffusion Models}.
\newblock \emph{arXiv:2411.18613}, 2024{\natexlab{b}}.

\bibitem[Wu et~al.(2024{\natexlab{c}})Wu, Mildenhall, Henzler, Park, Gao, Watson, Srinivasan, Verbin, Barron, Poole, and Holynski]{wu2024c}
R.~Wu, B.~Mildenhall, P.~Henzler, K.~Park, R.~Gao, D.~Watson, P.~P. Srinivasan, D.~Verbin, J.~T. Barron, B.~Poole, and A.~Holynski.
\newblock {ReconFusion: 3D Reconstruction with Diffusion Priors}.
\newblock In \emph{Computer Vision and Pattern Recognition Conference (CVPR)}, 2024{\natexlab{c}}.

\bibitem[Xu et~al.(2024{\natexlab{a}})Xu, Cheng, Gao, Wang, Gao, and Shan]{xu2024a}
J.~Xu, W.~Cheng, Y.~Gao, X.~Wang, S.~Gao, and Y.~Shan.
\newblock Instantmesh: Efficient 3d mesh generation from a single image with sparse-view large reconstruction models.
\newblock \emph{arXiv preprint arXiv:2404.07191}, 2024{\natexlab{a}}.

\bibitem[Xu et~al.(2024{\natexlab{b}})Xu, Tan, Luan, Bi, Wang, Li, Shi, Sunkavalli, Wetzstein, Xu, and Zhang]{xu2024}
Y.~Xu, H.~Tan, F.~Luan, S.~Bi, P.~Wang, J.~Li, Z.~Shi, K.~Sunkavalli, G.~Wetzstein, Z.~Xu, and K.~Zhang.
\newblock {DMV3D: Denoising Multi-View Diffusion using 3D Large Reconstruction Model}.
\newblock In \emph{International Conference on Learning Representations (ICLR)}, 2024{\natexlab{b}}.

\bibitem[Yang et~al.(2024{\natexlab{a}})Yang, Li, Fang, Liang, Xie, Zhang, Shen, and Tian]{yang2024a}
C.~Yang, S.~Li, J.~Fang, R.~Liang, L.~Xie, X.~Zhang, W.~Shen, and Q.~Tian.
\newblock {GaussianObject: High-Quality 3D Object Reconstruction from Four Views with Gaussian Splatting}.
\newblock In \emph{SIGGRAPH Asia}, 2024{\natexlab{a}}.

\bibitem[Yang et~al.(2023)Yang, Gao, Li, Gao, Wang, and Zheng]{yang2023}
J.~Yang, M.~Gao, Z.~Li, S.~Gao, F.~Wang, and F.~Zheng.
\newblock {Track Anything: Segment Anything Meets Videos}.
\newblock \emph{arXiv preprint arXiv:2304.11968}, 2023.

\bibitem[Yang et~al.(2025)Yang, Yu, Zeng, Lv, Ren, Lu, Lin, and Pang]{yang2025novel}
S.~Yang, W.~Yu, J.~Zeng, J.~Lv, K.~Ren, C.~Lu, D.~Lin, and J.~Pang.
\newblock Novel demonstration generation with gaussian splatting enables robust one-shot manipulation.
\newblock \emph{arXiv preprint arXiv:2504.13175}, 2025.

\bibitem[Yang et~al.(2024{\natexlab{b}})Yang, Gao, Zhou, Jiao, Zhang, and Jin]{yang2024}
Z.~Yang, X.~Gao, W.~Zhou, S.~Jiao, Y.~Zhang, and X.~Jin.
\newblock {Deformable 3D Gaussians for High-Fidelity Monocular Dynamic Scene Reconstruction}.
\newblock In \emph{Computer Vision and Pattern Recognition Conference (CVPR)}, 2024{\natexlab{b}}.

\bibitem[Yi et~al.(2024)Yi, Fang, Wang, Wu, Xie, Zhang, Liu, Tian, and Wang]{yi2024}
T.~Yi, J.~Fang, J.~Wang, G.~Wu, L.~Xie, X.~Zhang, W.~Liu, Q.~Tian, and X.~Wang.
\newblock {GaussianDreamer: Fast Generation from Text to 3D Gaussians by Bridging 2D and 3D Diffusion Models}.
\newblock In \emph{Computer Vision and Pattern Recognition Conference (CVPR)}, 2024.

\bibitem[Yoon et~al.(2020)Yoon, Kim, Gallo, Park, and Kautz]{yoon2020novel}
J.~S. Yoon, K.~Kim, O.~Gallo, H.~S. Park, and J.~Kautz.
\newblock Novel view synthesis of dynamic scenes with globally coherent depths from a monocular camera.
\newblock In \emph{Computer Vision and Pattern Recognition Conference (CVPR)}, 2020.

\bibitem[Yu et~al.(2024)Yu, Wang, Zhuang, Menapace, Siarohin, Cao, Jeni, Tulyakov, and Lee]{yu2024}
H.~Yu, C.~Wang, P.~Zhuang, W.~Menapace, A.~Siarohin, J.~Cao, L.~A. Jeni, S.~Tulyakov, and H.-Y. Lee.
\newblock {4Real: Towards Photorealistic 4D Scene Generation via Video Diffusion Models}.
\newblock In \emph{Conference on Neural Information Processing Systems}, 2024.

\bibitem[Zeng et~al.(2024)Zeng, Jiang, Zhu, Lu, Lin, Zhu, Hu, Cao, and Yao]{zeng2024}
Y.~Zeng, Y.~Jiang, S.~Zhu, Y.~Lu, Y.~Lin, H.~Zhu, W.~Hu, X.~Cao, and Y.~Yao.
\newblock {STAG4D: Spatial-Temporal Anchored Generative 4D Gaussians}.
\newblock In \emph{European Conference on Computer Vision (ECCV)}, 2024.

\bibitem[Zhang et~al.(2018)Zhang, Isola, Efros, Shechtman, and Wang]{zhang2018}
R.~Zhang, P.~Isola, A.~A. Efros, E.~Shechtman, and O.~Wang.
\newblock {The Unreasonable Effectiveness of Deep Features as a Perceptual Metric}.
\newblock In \emph{Computer Vision and Pattern Recognition Conference (CVPR)}, 2018.

\bibitem[Zhao et~al.(2024)Zhao, Colburn, Ma, Ángel Bautista, Susskind, and Schwing]{zhao2024}
X.~Zhao, A.~Colburn, F.~Ma, M.~Ángel Bautista, J.~M. Susskind, and A.~G. Schwing.
\newblock {Pseudo-Generalized Dynamic View Synthesis from a Video}.
\newblock In \emph{International Conference on Learning Representations (ICLR)}, 2024.

\bibitem[Zhou et~al.(2023)Zhou, Zhong, Shin, Lu, Yang, Markham, and Trigoni]{zhou2023}
K.~Zhou, J.-X. Zhong, S.~Shin, K.~Lu, Y.~Yang, A.~Markham, and N.~Trigoni.
\newblock {DynPoint: dynamic neural point for view synthesis}.
\newblock In \emph{Conference on Neural Information Processing Systems}, 2023.

\bibitem[Zhu et~al.(2024)Zhu, Liang, Chang, Deng, Lu, Yang, Zhang, and Zhang]{zhu2024}
R.~Zhu, Y.~Liang, H.~Chang, J.~Deng, J.~Lu, W.~Yang, T.~Zhang, and Y.~Zhang.
\newblock {MotionGS: Exploring Explicit Motion Guidance for Deformable 3D Gaussian Splatting}.
\newblock In \emph{Conference on Neural Information Processing Systems}, 2024.

\bibitem[Zou et~al.(2024)Zou, Yu, Guo, Li, Liang, Cao, and Zhang]{zou2024}
Z.-X. Zou, Z.~Yu, Y.-C. Guo, Y.~Li, D.~Liang, Y.-P. Cao, and S.-H. Zhang.
\newblock {Triplane Meets Gaussian Splatting: Fast and Generalizable Single-View 3D Reconstruction with Transformers}.
\newblock In \emph{Computer Vision and Pattern Recognition Conference (CVPR)}, 2024.

\end{thebibliography}
}

\newpage
\appendix
\maketitlesupplementary

\section{Additional Results}
In this section, we include additional qualitative and quantitative evaluation of \name{}.

\subsection{Further Qualitative Evaluation}
\begin{figure}
    \centering

    \includegraphics[width=0.24\linewidth]{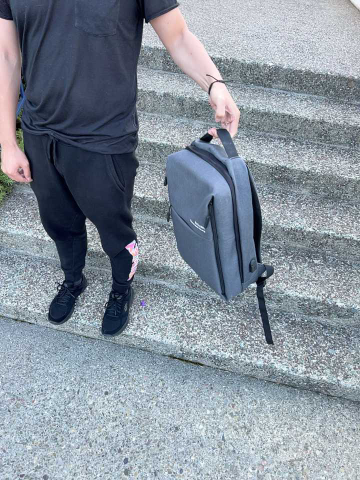}
    \includegraphics[width=0.24\linewidth]{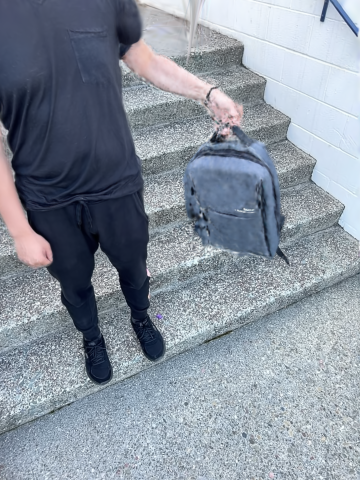}
    \includegraphics[width=0.24\linewidth]{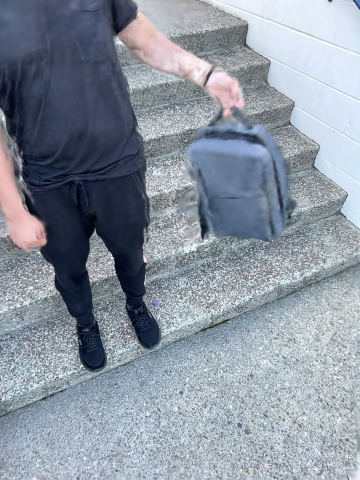}
    \includegraphics[width=0.24\linewidth]{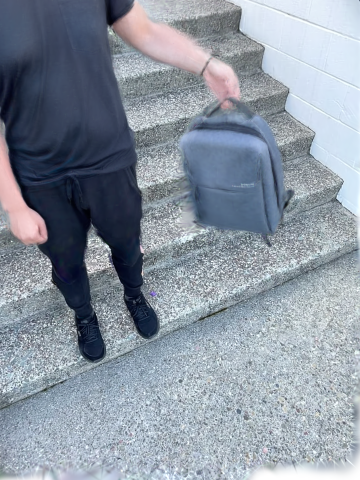}

    \includegraphics[width=0.24\linewidth]{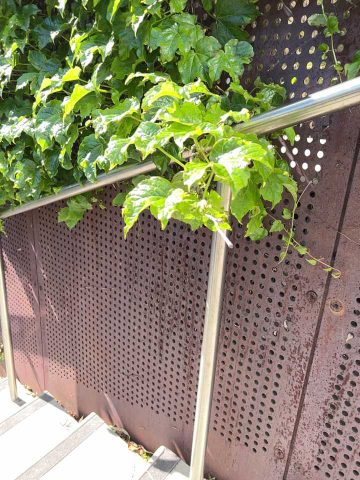}
    \includegraphics[width=0.24\linewidth]{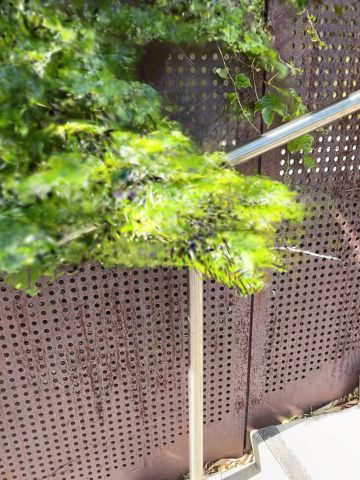}
    \includegraphics[width=0.24\linewidth]{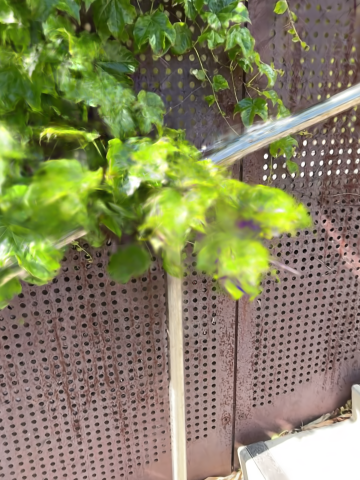}
    \includegraphics[width=0.24\linewidth]{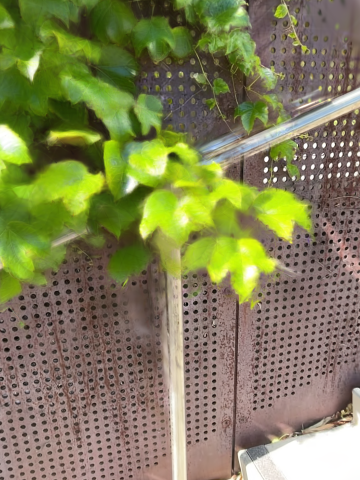}

    \includegraphics[width=0.24\linewidth]{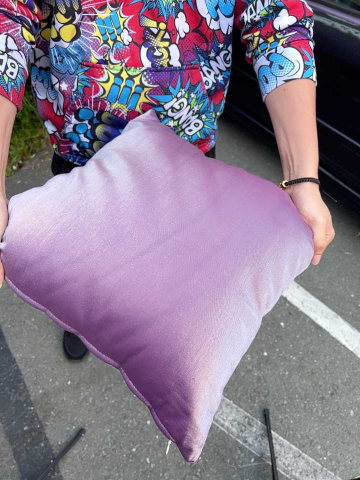}
    \includegraphics[width=0.24\linewidth]{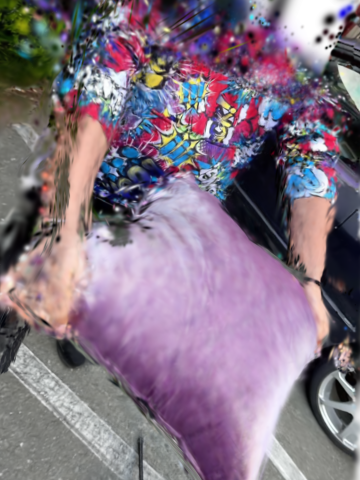}
    \includegraphics[width=0.24\linewidth]{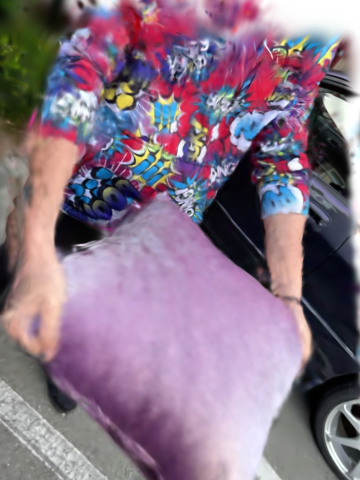}
    \includegraphics[width=0.24\linewidth]{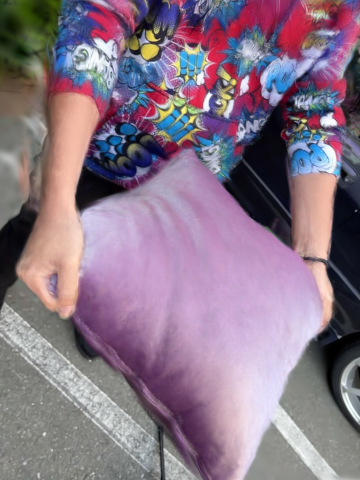}

    \includegraphics[width=0.24\linewidth]{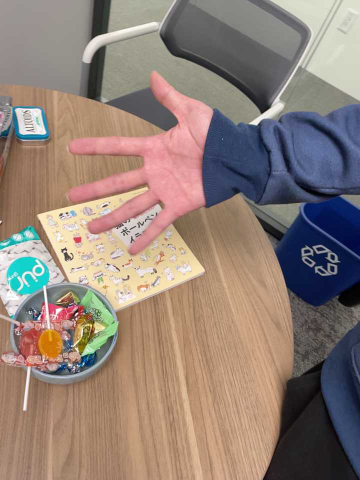}
    \includegraphics[width=0.24\linewidth]{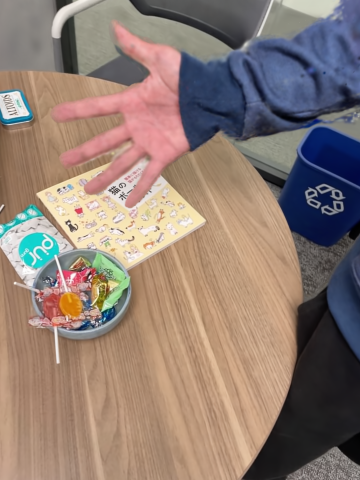}
    \includegraphics[width=0.24\linewidth]{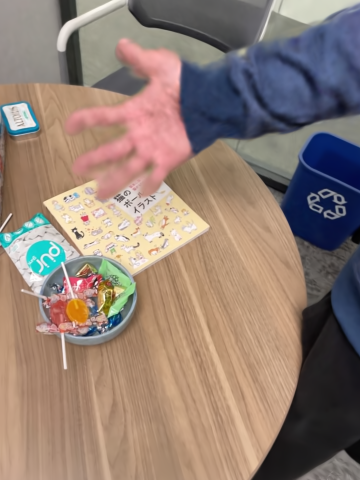}
    \includegraphics[width=0.24\linewidth]{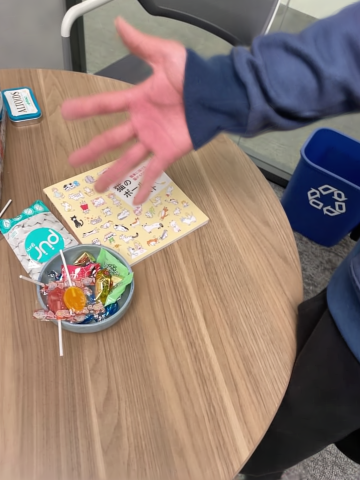}

    \includegraphics[width=0.24\linewidth]{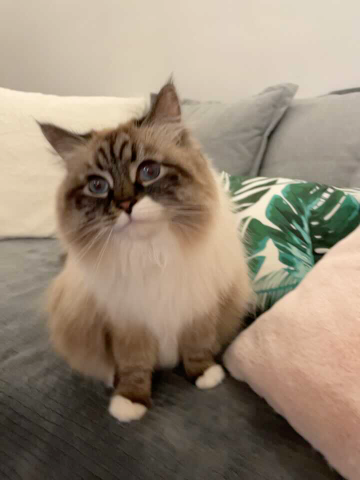}
    \includegraphics[width=0.24\linewidth]{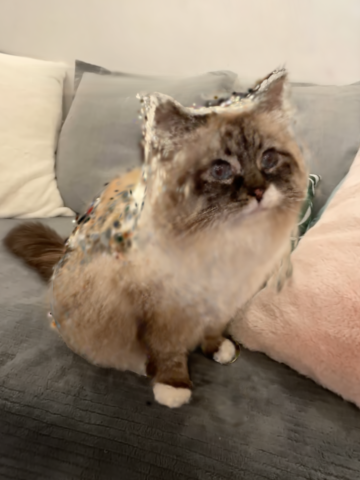}
    \includegraphics[width=0.24\linewidth]{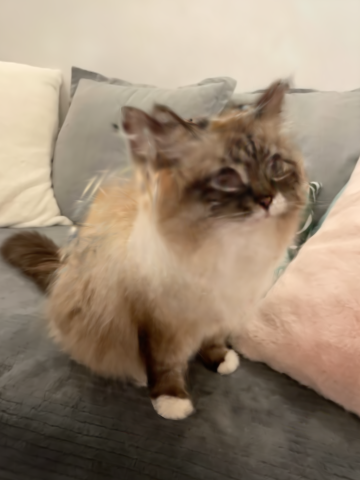}
    \includegraphics[width=0.24\linewidth]{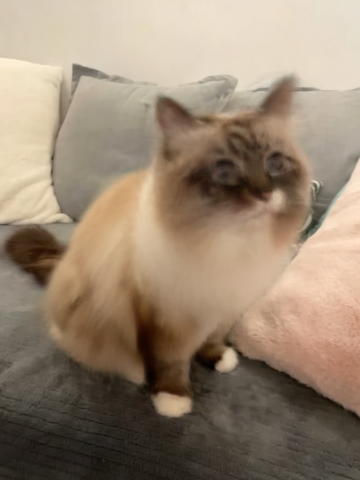}

    \small
    \begin{tabularx}{0.99\linewidth}{YYYY} 
     Input & Shape Of Motion\,\citep{wang2024} & MoSca\,\cite{lei2024} & Ours 
    \end{tabularx}
    \caption{Qualitative evaluation of our method against benchmark methods on the DyCheck qualitative example set.}
    \label{fig:quali-2}
\end{figure}

We present additional qualitative evaluation of \name{}~compared to MoSca and Shape Of Motion on the qualitative example scenes from the DyCheck dataset in Fig.~\ref{fig:quali-2}. Our results show consistently greater geometric consistency and visual quality compared to the other approaches.

\subsection{Per-Scene Results}

 We provide a detailed quantitative evaluation for every scene of the DyCheck dataset in Tables\,\ref{tab:supp_results_per_scene_halfres} and \ref{tab:supp_results_per_scene}, in half and full resolution respectively.  As in Tables\,\ref{tab:results-covis} and \ref{tab:results-dynamic}, we compute PSNR, SSIM and LPIPS on the co-visibility masked regions of the test views, which we denote with an \textit{-m} addendum to each metric,  as well as on the dynamic masked regions of the test views which we denote with a \textit{-D}. With a few exceptions (e.g. Apple, co-visibility), \name~is consistently the best performing method.

\begin{table}
\setlength{\tabcolsep}{4pt}
\centering
\caption{Per-scene quantitative evaluation of \name{} against state-of-the-art methods on the DyCheck dataset at half resolution. Best, second and third results are highlighted in red, orange and yellow respectively. }
\begin{tabularx}{\linewidth}{llZZZZZZ} 
     & {\small Method} & {\small PSNR-m\,\uparrowaligned} & {\small SSIM-m\,\uparrowaligned} & {\small LPIPS-m\,\downarrowaligned} & {\small PSNR-D\,\uparrowaligned} & {\small SSIM-D\,\uparrowaligned} & {\small LPIPS-D\,\downarrowaligned} \\ 
    \midrule
    \multirow{7}{*}{\STAB{\rotatebox[origin=c]{90}{Apple}}} & T-NeRF\,\citep{gao2022} & 17.43 & 0.7285 & 0.5081 & 13.63 & 0.9433 & 0.3108 \\ 
      & Nerfies\,\citep{park2021b}& 17.54 & 0.7505 & 0.4785 & 13.63 & \cellthird0.9437 & 0.3321 \\ 
      & HyperNeRF\,\citep{park2021a} & 17.64 & 0.7433 & 0.4775 & 13.36 & 0.9417 & 0.3362\\
    & Gaussian Marbles\,\cite{stearns2024} & 17.90 & 0.7328 & 0.4716 & 14.56 & 0.9370 & 0.3229 \\ 
    & SoM\,\cite{wang2024} & \cellthird18.95 & \cellbest0.8111 & \cellbest0.2917 & \cellthird14.88 & 0.9419 & \cellthird0.3016\\
    & MoSca\,\cite{lei2024} & \cellbest19.40 & \cellsecond0.8074 & \cellthird0.3392 & \cellsecond16.97 & \cellsecond0.9580 & \cellsecond0.2176\\
    & Ours & \cellsecond19.18 & \cellthird0.8008 & \cellsecond0.3149 & \cellbest17.75 & \cellbest0.9616 &\cellbest 0.1893\\
    \cmidrule(l){2-8}
    \multirow{7}{*}{\STAB{\rotatebox[origin=c]{90}{Block}}}  & T-NeRF\,\citep{gao2022} & 17.52 & 0.6688 & 0.3460 & 14.07 & 0.8591 & \cellthird0.3474\\ 
      & Nerfies\,\citep{park2021b}& 16.61 & 0.6393 & 0.3893 & 13.31 & 0.8433 & 0.4068 \\ 
      & HyperNeRF\,\citep{park2021a} & 17.54 & \cellthird0.6702 & \cellthird0.3312 & 13.73 & 0.8525 & 0.3483\\
    & Gaussian Marbles\,\cite{stearns2024} & 16.95 & 0.6509 & 0.3788 & 13.81 & 0.8480 & 0.3750\\ 
    & SoM\,\cite{wang2024} & \cellthird17.99 & 0.6634 & \cellsecond0.2608 & \cellthird15.23 & \cellthird0.8596 & \cellsecond0.3378\\
    & MoSca\,\cite{lei2024} & \cellsecond18.11 & \cellsecond0.6801 & 0.3416 & \cellsecond15.32 & \cellsecond0.8699 & 0.3499\\
    & Ours & \cellbest 18.91 & \cellbest0.6901 & \cellbest0.2168 & \cellbest15.93 & \cellbest0.8864 & \cellbest0.3052\\
    \cmidrule(l){2-8}
    \multirow{7}{*}{\STAB{\rotatebox[origin=c]{90}{Paper}}}  & T-NeRF\,\citep{gao2022} & 17.55 & 0.3672 & 0.2577 & 12.50 & 0.9730 & 0.3149\\ 
      & Nerfies\,\citep{park2021b}& 17.34 & 0.3783 & 0.2111 & 10.84 & 0.9710 & 0.3929\\ 
      & HyperNeRF\,\citep{park2021a} & 17.38 & 0.3819 & 0.2086 & 10.64 & 0.9706 & 0.4040\\
    & Gaussian Marbles\,\cite{stearns2024} & 16.62 & 0.3219 & 0.3517 & 11.68 & 0.9710 & \cellsecond0.2680\\ 
    & SoM\,\cite{wang2024}& \cellthird20.85 & \cellthird0.6725 & \cellsecond0.1536 & \cellthird12.90 & \cellthird0.9738 & \cellbest0.2634 \\
    & MoSca\,\cite{lei2024}& \cellsecond22.24 & \cellsecond0.7450 & \cellthird0.1617 & \cellsecond14.32 &\cellsecond 0.9789 & \cellthird0.2832 \\
    & Ours & \cellbest 22.48 & \cellbest0.7516 & \cellbest0.1287 & \cellbest15.58 & \cellbest0.9807 & 0.3080\\
    \cmidrule(l){2-8}
    \multirow{7}{*}{\STAB{\rotatebox[origin=c]{90}{Space-out}}}  & T-NeRF\,\citep{gao2022}& 17.71 & 0.5914 & 0.3768 & 14.52 & 0.8620 & 0.3649 \\ 
      & Nerfies\,\citep{park2021b}& 17.79 & 0.6217 & 0.3032 & 13.85 & 0.8578 & 0.3407 \\ 
      & HyperNeRF\,\citep{park2021a} & 17.93 & 0.6054 & 0.3203 & 14.26 & 0.8597 & 0.3379\\
    & Gaussian Marbles\,\cite{stearns2024} & 15.32 & 0.5512 & 0.4235 & 11.15 & 0.8488 & 0.4278\\ 
    & SoM\,\cite{wang2024} & \cellthird19.64 & \cellthird0.6313 & \cellsecond0.2374 & \cellthird15.72 & \cellthird0.8805 & \cellthird0.2429\\
    & MoSca\,\cite{lei2024} & \cellsecond20.35 & \cellsecond0.6582 & \cellthird0.2702 & \cellsecond17.57 & \cellsecond0.8985 & \cellbest0.2073\\
    & Ours & \cellbest21.58 & \cellbest0.6890 & \cellbest0.2010 & \cellbest18.77 & \cellbest0.9091 & \cellsecond0.2306\\
    \cmidrule(l){2-8}
    \multirow{7}{*}{\STAB{\rotatebox[origin=c]{90}{Spin}}}  & T-NeRF\,\citep{gao2022}& 19.16 & 0.5672 & 0.4427 & 15.65 & 0.9090 & 0.3394 \\ 
      & Nerfies\,\citep{park2021b} & 18.38 & 0.5846 & 0.3087 & 14.11 & 0.8899 & 0.3868\\ 
      & HyperNeRF\,\citep{park2021a} & 19.20 & 0.5614 & 0.3254 & 15.65 & 0.9060 & 0.3230\\
    & Gaussian Marbles\,\cite{stearns2024} & 18.31 & 0.5461 & 0.3558 & 15.68 & 0.8963 & 0.3950\\ 
    & SoM\,\cite{wang2024}& \cellthird21.05 & \cellbest0.7798 & \cellbest0.1698 & \cellthird18.44 & \cellthird0.9180 & \cellthird0.2815 \\
    & MoSca\,\cite{lei2024} & \cellsecond21.06 & \cellthird0.7100 & \cellthird0.2084 & \cellsecond18.85 & \cellsecond0.9240 & \cellsecond0.2688\\
    & Ours & \cellbest21.28 & \cellsecond0.7209 & \cellsecond0.1853 & \cellbest19.37 & \cellbest0.9327 & \cellbest0.2566\\
    \cmidrule(l){2-8}
    \multirow{7}{*}{\STAB{\rotatebox[origin=c]{90}{Teddy}}}  & T-NeRF\,\citep{gao2022} & 13.71 & \cellthird0.5695 & 0.4286 & 13.49 & \cellthird0.6198 & 0.3730\\ 
      & Nerfies\,\citep{park2021b} & 13.65 & 0.5572 & 0.3716 & 13.33 & 0.6050 & 0.3280\\ 
      & HyperNeRF\,\citep{park2021a} & 13.97 & 0.5678 & 0.3498 & 13.65 & 0.6168 & \cellsecond0.3028\\
    & Gaussian Marbles\,\cite{stearns2024} & 13.65 & 0.5432 & 0.4369 & 13.29 & 0.5972 & 0.3961\\ 
    & SoM\,\cite{wang2024}& \cellthird14.00 & 0.5462 & \cellsecond0.3399 & \cellthird13.68 & 0.5998 & 0.3335 \\
    & MoSca\,\cite{lei2024} & \cellsecond15.09 & \cellsecond0.6133 & \cellthird0.3587 & \cellsecond14.84 & \cellsecond0.6679 & \cellthird0.3146\\
    & Ours & \cellbest15.97 & \cellbest0.6416 & \cellbest0.3096 & \cellbest15.62 & \cellbest0.6844 & \cellbest0.2901\\
    \cmidrule(l){2-8}
    \multirow{7}{*}{\STAB{\rotatebox[origin=c]{90}{Wheel}}}  & T-NeRF\,\citep{gao2022} & 15.65 & 0.5481 & 0.2925 & \cellbest13.19 & 0.8162 & 0.3937\\ 
      & Nerfies\,\citep{park2021b} & 13.82 & 0.4580 & 0.3097 & 11.15 & 0.7870 & 0.4805\\ 
      & HyperNeRF\,\citep{park2021a} & 13.99 & 0.4550 & 0.3102 & 11.60 & 0.7913 & 0.4385\\
    & Gaussian Marbles\,\cite{stearns2024} & 16.02 & 0.5416 & 0.3398 & 9.99 & 0.8175 & 0.3926\\ 
    & SoM\,\cite{wang2024}& \cellthird17.86 & \cellthird0.6695 & \cellsecond0.2144 & \cellsecond12.75 & \cellsecond0.8338 & \cellbest0.3446 \\
    & MoSca\,\cite{lei2024} & \cellsecond17.95 & \cellsecond0.6852 & \cellthird0.2309 & 11.57 & \cellthird0.8310 & \cellthird0.3918\\
    & Ours & \cellbest18.46 & \cellbest0.6943 & \cellbest0.2058 & \cellthird12.23 & \cellbest0.8402 & \cellsecond0.3754\\
    \bottomrule
\end{tabularx}
\label{tab:supp_results_per_scene_halfres}
\end{table}

\begin{table}
\setlength{\tabcolsep}{4pt}
\centering
\caption{Per-scene quantitative evaluation of \name{} against state-of-the-art methods on the DyCheck dataset at full resolution. Best, second and third results are highlighted in red, orange and yellow respectively. }
\begin{tabularx}{\linewidth}{llZZZZZZ} 
     & {\small Method} & {\small PSNR-m\,\uparrowaligned} & {\small SSIM-m\,\uparrowaligned} & {\small LPIPS-m\,\downarrowaligned} & {\small PSNR-D\,\uparrowaligned} & {\small SSIM-D\,\uparrowaligned} & {\small LPIPS-D\,\downarrowaligned} \\ 
    \midrule
    \multirow{4}{*}{\STAB{\rotatebox[origin=c]{90}{Apple}}} & Gaussian Marbles\,\cite{stearns2024} &16.84 & 0.7022 & 0.6849 & 14.65 & 0.9495 & 0.4281\\ 
    & SoM\,\cite{wang2024} & \cellthird17.74 & \cellbest0.7535 & \cellbest0.4946 & \cellthird14.88 & \cellthird0.9540 & \cellthird0.4036 \\
    & MoSca\,\cite{lei2024}& \cellbest18.19 & \cellsecond0.7486 & \cellthird0.5651 & \cellsecond17.21 & \cellsecond0.9681 & \cellsecond0.3073 \\
    & Ours & \cellsecond18.02 & \cellthird0.7466 & \cellsecond0.5359 & \cellbest18.07 & \cellbest0.9694 & \cellbest0.2559\\
    \cmidrule(l){2-8}
    \multirow{4}{*}{\STAB{\rotatebox[origin=c]{90}{Block}}} & Gaussian Marbles\,\cite{stearns2024} & 16.50 & 0.6492 & 0.5065& 13.49 & 0.8660 & 0.4987 \\ 
    & SoM\,\cite{wang2024}& \cellthird17.42 & \cellthird0.6566 & \cellbest 0.3879 & \cellthird14.60 & \cellthird0.8759 & \cellsecond0.4667 \\
    & MoSca\,\cite{lei2024} & \cellsecond17.56 & \cellsecond0.6710 & \cellthird0.4658 & \cellsecond14.97 & \cellsecond0.8848 & \cellthird0.4808\\
    & Ours &\cellbest 18.43 & \cellbest0.6722 & \cellsecond0.3932 & \cellbest15.54 & \cellbest0.8898 & \cellbest0.4100 \\
    \cmidrule(l){2-8}
    \multirow{4}{*}{\STAB{\rotatebox[origin=c]{90}{Paper}}} & Gaussian Marbles\,\cite{stearns2024} & 15.96 & 0.2959 & 0.5778 & 11.30 & 0.9722 & \cellthird0.5094\\ 
    & SoM\,\cite{wang2024}& \cellthird19.65 & \cellthird0.5518 & \cellsecond0.2035 & \cellthird13.03 & \cellthird0.9737 & \cellbest0.4861 \\
    & MoSca\,\cite{lei2024}& \cellsecond20.82 & \cellsecond0.6289 & \cellthird0.2412 & \cellsecond14.04 & \cellsecond0.9770 & 0.5498 \\
    & Ours & \cellbest21.06 & \cellbest0.6477 & \cellbest0.1923 & \cellbest14.99 & \cellbest0.9777 & \cellsecond0.5003\\
    \cmidrule(l){2-8}
    \multirow{4}{*}{\STAB{\rotatebox[origin=c]{90}{Space-out}}} & Gaussian Marbles\,\cite{stearns2024} & 15.19 & 0.5603 & 0.5434 & 11.07 & 0.8671 & 0.5334\\ 
    & SoM\,\cite{wang2024} & \cellthird19.54 & \cellthird0.6178 & \cellsecond0.3667 & \cellthird16.11 & \cellthird0.8939 & \cellthird0.3363\\
    & MoSca\,\cite{lei2024} & \cellsecond19.93 & \cellsecond0.6280 & \cellthird0.4060 & \cellsecond17.17 & \cellsecond0.8988 & \cellbest0.3207\\
    & Ours & \cellbest21.15 & \cellbest0.6443 & \cellbest0.3278 & \cellbest18.56 & \cellbest0.9079 & \cellsecond0.3356\\
    \cmidrule(l){2-8}
    \multirow{4}{*}{\STAB{\rotatebox[origin=c]{90}{Spin}}} & Gaussian Marbles\,\cite{stearns2024} & 17.84 & 0.5154 & 0.4784 & 15.79 & 0.9143 & 0.4420 \\ 
    & SoM\,\cite{wang2024} & \cellthird20.57 & \cellbest0.7323 & \cellbest0.2663 & \cellsecond18.65 & \cellthird0.9339 & \cellsecond0.3535\\
    & MoSca\,\cite{lei2024} & \cellsecond20.61 & \cellthird0.6769 & \cellthird0.3108 & \cellthird18.50 & \cellsecond0.9371 & \cellthird0.3657\\
    & Ours & \cellbest20.69 & \cellsecond0.6851 & \cellsecond0.2868 & \cellbest19.37 &\cellbest 0.9445 & \cellbest0.3094\\
    \cmidrule(l){2-8}
    \multirow{4}{*}{\STAB{\rotatebox[origin=c]{90}{Teddy}}} & Gaussian Marbles\,\cite{stearns2024} & 13.01 & 0.5496 & 0.6559 & 13.05 & \cellthird0.6269 & 0.5815  \\ 
    & SoM\,\cite{wang2024} & \cellthird13.58 & \cellthird0.5518 & \cellsecond0.5377 & \cellthird13.40 & 0.6251 & \cellsecond0.4928 \\
    & MoSca\,\cite{lei2024}& \cellsecond14.52 & \cellsecond0.5925 & \cellthird0.5777 & \cellsecond14.41 & \cellsecond0.6692 & \cellthird0.4986 \\
    & Ours & \cellbest15.63 & \cellbest0.6238 & \cellbest0.4786 & \cellbest15.46 &\cellbest 0.6892 &\cellbest0.4133\\
    \cmidrule(l){2-8}
    \multirow{4}{*}{\STAB{\rotatebox[origin=c]{90}{Wheel}}} & Gaussian Marbles\,\cite{stearns2024} & 15.55 & 0.5310 & 0.5295 & 9.89 & 0.8291 & \cellthird0.5473\\ 
    & SoM\,\cite{wang2024} & \cellthird17.38 & \cellthird0.6318 & \cellsecond0.3456 & \cellbest13.05 & \cellthird0.8393 & \cellbest0.5037\\
    & MoSca\,\cite{lei2024}& \cellsecond17.49 & \cellsecond0.6460 & \cellthird0.3684 & \cellthird11.42 & \cellsecond0.8395 & 0.5665 \\
    & Ours & \cellbest18.00 & \cellbest0.6505 & \cellbest0.3215 & \cellsecond12.21 & \cellbest0.8469 & \cellsecond0.5203\\
    \bottomrule
\end{tabularx}
\label{tab:supp_results_per_scene}
\end{table}

\section{Implementation Details}

In this section, we provide any implementation details not included in the main manuscript.

\subsection{Monocular Reconstruction}

We implement the monocular reconstruction step directly as MoSca\,\cite{lei2024}, keeping the original hyperparameters intact. We substitute dynamic masks estimated from epipolar error by masks obtained from Track Anything\,\cite{yang2023}.

\subsection{Personalised Diffusion Model}

We train our personalised diffusion model with a Dreambooth\,\cite{ruiz2023} approach implemented in the $\mathtt{diffusers}$\footnote{\url{https://huggingface.co/docs/diffusers}} library as a LoRA fine-tuning process. We use the default implementation of the SDXL model with default parameters. We change the resolution to match our input resolution (720x960). Similarly, we change the number of training iterations from the default 500 to 5000, in response to the default model being suitable for personalisation with a smaller number of images (5-40), as opposed to our inputs (ranging above 400).

\subsection{Camera Sampling}

To obtain a set of varying samples for multi-view supervision, we propose a camera sampling strategy based on extreme poses within the input trajectory. 

Given the set of input camera poses (position and orientation), we calculate a mean camera pose. Then, we establish a sphere approximating the surface established by the input trajectory, assuming that a target dynamic object is being tracked by the recording. Finally, we select two views in the input trajectory that, when projected on the sphere, are characterised by the largest longitudinal displacement. These constitute the extreme camera poses.

Thereafter, for each time step spanning the whole time range of the input video, we sample the following new cameras:
\begin{itemize}    
    \item Two random camera poses from the input trajectory are selected, and a new camera pose is calculated as their mean, and random noise is added.  Total cameras: 4

    \item For each of the two extreme views, a random camera pose from the input trajectory is selected, and a new camera pose is calculated as their weighted average, and random noise is added, with the weight increasing towards the extreme views. Total cameras: 12   

    \item The extreme camera views. Total cameras: 2
    
\end{itemize}
This constitutes our set of 18 new training cameras for each timestep of the input video $c_m \in \mathcal{C}_{sample}$.

\subsection{Multi-View Sample Enhancement}

Having sampled a set of new trajectories, we render them with the previously trained monocular reconstruction model, in such way we obtain a set of degraded images $\{ R_{m,t}\}$. To perform the enhancement as described in Section \ref{sec:diffusion}, we utilise the Image2Image translation approach as implemented in $\mathtt{diffusers}$.

\subsection{Diffusion-Aware Reconstruction}
We increase the total number of iterations from 8000 to 40000 in order to train on the additional generated data. During optimisation, we run two separate forward and backward passes, the first for sampled camera pose optimisation and the second for optimising the Gaussians and input camera poses. At each iteration, we randomly select two of the sampled camera poses which correspond to the same time step as the input camera. During the first pass, we render the images and compute the mean of the camera losses $\mathcal{L}_{cam}$ for both of the sampled cameras and update only the sampled camera poses $\mathcal{C}_{sample}$. During the second pass, we re-render the images using the updated camera poses and compute the dynamic loss $\mathcal{L}_{dyn}$ using the dynamic region masks. This loss is added to the existing monocular losses and is used to update the input camera poses $\mathcal{C}_{inp}$ and the Gaussians $\mathcal{G}$.

\end{document}